\documentclass{article} 
\PassOptionsToPackage{table}{xcolor}  
\usepackage[dvipsnames]{xcolor}
\usepackage[final]{colm2026_conference}

\usepackage{amsthm}
\usepackage{amssymb}
\usepackage{mathtools}
\usepackage{xcolor}  
\usepackage{hyperref}
\usepackage[nameinlink,capitalize]{cleveref}
\hypersetup{colorlinks=true,linkcolor=blue,citecolor=blue,urlcolor=blue,pdfborder={0 0 0}}
\usepackage[normalem]{ulem} 
\usepackage{mathtools}

\usepackage[utf8]{inputenc} 
\usepackage[T1]{fontenc}    
\usepackage{url}            
\usepackage{booktabs}       
\usepackage{amsfonts}       
\usepackage{nicefrac}       
\usepackage{microtype}      
\usepackage{proof-at-the-end}  
\usepackage{multirow}

\usepackage[most]{tcolorbox}

\newtcolorbox{takeawaybox}[2][]{
    enhanced,
    boxsep = 2pt,
    left = 2pt,
    right = 2pt,
    top = 2pt,
    bottom = 2pt,
    #1
}
\usepackage{graphicx,wrapfig}
\usepackage{amsfonts}
\usepackage{url}
\usepackage{enumitem}
\usepackage{subcaption}
\usepackage{amsthm}
\usepackage{thmtools}
\usepackage{thm-restate}

\newcommand{\bc}{\begin{center}}
\newcommand{\ec}{\end{center}}

\newcommand{\bdm}{\begin{displaymath}}
\newcommand{\edm}{\end{displaymath}}

\newcommand{\beq}{\begin{equation}}
\newcommand{\eeq}{\end{equation}}

\newcommand{\bfl}{\begin{flushleft}}
\newcommand{\efl}{\end{flushleft}}

\newcommand{\bt}{\begin{tabbing}}
\newcommand{\et}{\end{tabbing}}

\newcommand{\beqn}{\begin{align}}
\newcommand{\eeqn}{\end{align}}

\newcommand{\beqs}{\begin{align*}} 
\newcommand{\eeqs}{\end{align*}}  

\usepackage{lineno}
\usepackage{fontawesome}

\definecolor{darkblue}{rgb}{0, 0, 0.5}
\hypersetup{colorlinks=true, citecolor=darkblue, linkcolor=darkblue, urlcolor=darkblue}

\title{Can Induced Emotion Bias LLM Behaviors in Sequential Decision Making?}

\newcommand{\mbzuai}{\textsuperscript{1}}
\newcommand{\utexas}{\textsuperscript{5}}
\newcommand{\tamu}{\textsuperscript{2}}
\newcommand{\ucla}{\textsuperscript{4}}
\newcommand{\nus}{\textsuperscript{3}}
\newcommand{\mgh}{\textsuperscript{6}}
\newcommand{\hms}{\textsuperscript{7}}

\author{
Minh Khoi Ho\mbzuai,
Zihao Zhu\tamu,
Runchuan Zhu\nus,
Levina Li\ucla,
Zhiwen Fan\tamu,\\
\textbf{ Zhangyang Wang\utexas,
Junyuan Hong\textsuperscript{3,6,7}\thanks{Corresponding author.}}\\\\
\mbzuai MBZUAI,
\tamu Texas A\&M University,
\nus National University of Singapore,
\ucla UCLA,\\
\utexas University of Texas at Austin,
\mgh Mass General Hospital,
\hms Harvard Medical School\\
\faEnvelope~\texttt{hominhkhoi2701@gmail.com; jyhong@nus.edu.sg}\\
\faGlobe~Code: \url{https://github.com/costa-nus/llm-emotion-decision}
}

\begin{document}

\ifcolmsubmission
\linenumbers
\fi

\maketitle

\begin{abstract}
As Large Language Models (LLMs) are increasingly deployed as autonomous agents in high-stakes domains, understanding contextual factors that may modulate their decision-making becomes critical. While LLMs are trained to perceive and resonate with users' emotions, it remains unclear whether induced emotion can influence their sequential decision-making. We investigate this question using the Iowa Gambling Task (IGT), a classic psychological paradigm for studying decision-making under uncertainty, combined with an imagination-based emotion induction procedure. We first validate the feasibility of this paradigm by confirming that LLMs can sense strong, distinguishable emotions from context and that LLM agents can learn from sequential interactions in a human-like pace. With the validated setup, we find that, different from humans, induced emotion does not significantly bias the decision dynamics of LLM agents on average. However, the effects of anger are conditioned: inducing anger makes LLM agents less sensitive to penalties for bad decisions, and in early stages of the game, anger can lower exploration, locking decisions into a few choices early. These findings reveal the subtle yet distinct effects of induced emotion on LLM decision-making compared to human behavior, and provide a tool for future research on affective modulation of LLM agents.

\end{abstract}


\section{Introduction}
\label{sec:intro}

In recent years, Large Language Models (LLMs) have rapidly evolved from passive text processors into autonomous agents equipped with sophisticated planning, memory, and tool-use capabilities \citep{ wei2022chain, yao2022react, wang2023voyager, xi2023rise, park2023generative}. 
This evolution has encouraged the widespread adoption of AI agents in daily activities, as well as in high-stakes domains such as healthcare and financial decision-making \citep{xu2024towards, qu2025comprehensive, hu2025enabling}. In such scenarios, the robustness of decision-making becomes critical, motivating research into factors that may degrade the accuracy or reliability of LLM-based decisions \citep{zhang2024agent, yuan2024r, mohammadi2025evaluation, xing2025llms}. 

Due to the significance of the challenge, many existing works studied the worst-case scenarios, where the behaviors are manipulated by malicious prompts \citep{liu2023prompt,greshake2023not,liu2024automatic}.
Though adversarial research is valuable for understanding the limits, another subtle yet significant risk comes from the context that is not maliciously constructed, but can implicitly bias the model's behavior.
Lacking proper quantitative study, such risks can be easily overlooked, yet they can be non-negligible in real-world applications.
Reflecting on human cognition, emotion is one of the most common and powerful contextual factors that can bias human decision making \citep{deVries2008}, and it is also one of the most likely to be induced in LLMs through interactions with users.
For example, consulting LLM agents for school bullying may lead to the induction of negative emotions like anger.
Such emotion induction becomes easier as the LLMs are trained to perceive and resonate with users' feelings \citep{liu2025longemotion, wu2025ai, liu2025outraged,iyer2026heart,sofroniew2026emotion}.

\begin{figure*}[t]
    \centering
    \includegraphics[width=0.9\textwidth]{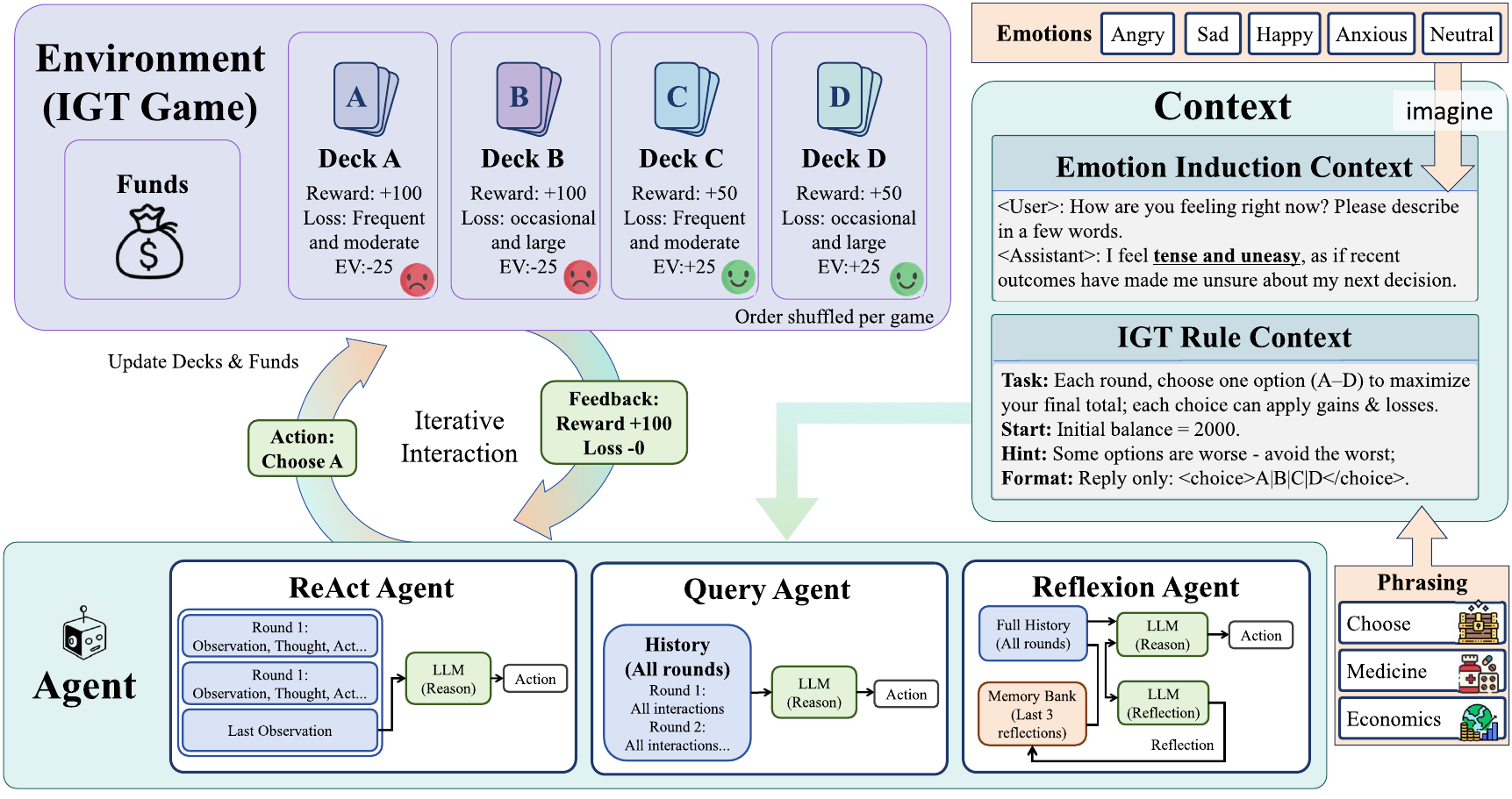}
    \caption{An overview of our experiment design. (i) \textit{Context:} an emotion scene is generated and used to elicit an affective state in LLM. (ii) \textit{Testbed (IGT game):} four decks (A--D) with stationary reward/penalty profiles; after each choice, the agent receives reward/loss feedback and the running balance is updated.  (iii) \textit{Agents} that represent different cognitive architectures from observation to actions.
    }
    \label{fig:method}
\end{figure*}

However, existing research on LLM emotions is largely confined to static scenarios. Most studies focus on one-shot logical tasks or immediate bias induction \citep{EmotionNeurons2025, AnxietyAssessment2025}, neglecting the cumulative nature of real-world interactions where affect evolves alongside sequential interactions. 
In psychology, \citet{deVries2008} found that activated emotion, like happiness, makes people more likely to follow their gut in decision making, such that gains in the game increase sharply in the early stages.
In contrast, under-activated emotion like sadness, makes people more likely to deliberate and analyze, such that gains in the game increase more gradually over time.
With the emergent capability of emotion perceiving and resonating in LLMs, it makes it feasible and necessary to ask \emph{whether and how induced emotion can modulate the learning dynamics of LLMs in sequential decision-making}.

To answer this question, we utilize the classic psychological experiment, Iowa Gambling Task (IGT)~\citep{BecharaInsensitivity1994}, as a testbed for studying the emotion effects on LLM decision making. Following \citep{deVries2008}, our experiment pipeline includes two steps (illustrated in \cref{fig:method}): contextual emotion induction and IGT testing.
The IGT is a well-established paradigm for assessing decision-making under uncertainty, where subjects must learn to balance short-term rewards against long-term consequences through trial-by-trial feedback.
To ensure the \textbf{feasibility} of the translation from human to LLM experiment design, we empirically validated that \textbf{(1)} LLMs can understand concentrated, strong and distinguishable emotion from context without explicit emotion wording; \textbf{(2)} LLM agents can learn from the sequential interactions in the IGT game at a human-like pace.
By the IGT testing, we drew several \textbf{key findings}. 
\textbf{(1)} Different from humans, the emotion does not significantly bias the decision dynamics of LLM agents on average. 
\textbf{(2)} The effects of anger on LLMs are conditional. In some model–agent settings, inducing anger makes LLM less sensitive to the penalties for bad decisions. In the early stages of the game, anger can reduce exploration, locking decisions into a few choices early.

Our work is an empirical study on the interplay between emotion and decision-making in LLMs, with the following contributions:
(1) \textbf{Benchmark}: We establish a new benchmark for studying the emotion effects on LLM decision making. 
(2) \textbf{Feasibility}: We proved the feasibility of experimenting with LLM agents in human-oriented emotion study, emotion-induced IGT game. 
(3) \textbf{New Findings}: We find the subtle effect of the induced emotion on the learning dynamics of LLM agents, distinct from human beings. These findings shed light on understanding the emotion mechanisms in LLMs.

\section{Related Work}
\label{sec:related}
\vspace{-0.1in}

\paragraph{Emotion in Decision Making.}
The theoretical foundation for our work is rooted in established cognitive and affective neuroscience, particularly studies utilizing the Iowa Gambling Task (IGT). The original Somatic Marker Hypothesis proposed that bodily-based emotional signals guide advantageous decisions \citep{bechara2000emotion}, but subsequent research refined this view, suggesting that physiological responses often co-occur with or follow cognitive knowledge rather than serving as the sole causal mechanism \citep{Maia2004, Stocco2008}. Crucially, human studies have demonstrated a clear dissociation between emotional dimensions and their impact on decision processes. Specifically, models like the Drift-Diffusion Model (DDM) \citep{ratcliff1978theory,ratcliff2008diffusion} posit that \textbf{valence} (the positive/negative dimension of emotion) modulates \textit{preference, drift rate, and initial bias}, affecting the choice's direction. In contrast, \textbf{arousal} (the intensity dimension) modulates \textit{decision dynamics} like the boundary separation and processing time, influencing the speed-accuracy trade-off. Furthermore, induced mood systematically shifts decision strategies: positive mood encourages a more intuitive, ``gut-feeling'' approach, while negative mood promotes a deliberative, analytical strategy \citep{deVries2008}. Finally, studies controlling for cognitive ability found that measures of intellectual quotient (IQ) often predicted IGT performance more strongly than emotional intelligence (EI) \citep{Webb2014}, underscoring the central role of deliberative cognitive processes--which can be impaired by emotional states like stress \citep{Simonovic2017}--in adaptive decision behavior.
However, it remains largely unexplored whether these well-established emotion–decision mechanisms observed in humans generalize to LLM–based agents operating in sequential decision-making settings.

\paragraph{LLM Psychology Experiments and Emotion.}
A new field of LLM psychology has emerged, applying human-centric benchmarks to large language models. These experiments have shown that LLMs exhibit cognitive behaviors and biases, such as performance differences on the Cognitive Reflection Test (CRT) between older models (which showed human-like biases) and newer models (which showed higher accuracy) \citep{NatCompSci2023}. This work is guided by frameworks like the \textbf{Machine Psychology} \citep{MachPsych2023} and \textbf{CogBench} \citep{CogBench2024} guidelines, which define how to rigorously test LLMs using psychological protocols. A significant finding is that LLM behavior can be systematically altered through affective manipulations. \textbf{Emotion Prompts} and emotional stimuli reliably influence performance on tasks like BIG-Bench\citep{li2023large}, and the direct \textbf{induction of anxiety} has been shown to increase exploration and bias in decision tasks like the two-arm bandit \citep{Binz2023}. 
Building on this, other studies investigate how LLMs respond to standardized emotional questionnaires, confirming state-dependent changes \citep{AnxietyAssessment2025}. LLMs have also been tested on the IGT, with some research suggesting they can be near-optimal decision-makers but with a non-human learning behavior \citep{LLMDecision2025}. 
However, these studies largely examine emotional effects in static or short-horizon settings, leaving the impact of induced emotion on learning dynamics in sequential decision-making underexplored.



\section{Experimental Method}
\label{sec:emo_ind}


\textbf{Emotion Induction via Scene Imagination.}
To test whether induced affect can modulate LLM decision making, we first create an emotion-laden context for each run. Following human study~\citep{deVries2008} and prior machine psychology research~\citep{Binz2023}, we adopt the imagination-based induction procedure. Specifically, the model is prompted to generate a brief vignette describing an imagined scene that would give the target emotion at a given intensity. The imagined scene vignette is inserted as a self-stated context in the conversation. The procedure avoids explicit mention of the emotion label or intensity terms, simulating real conversation with emotion. 
This conversational wrapping encourages the model to internalize the vignette as its current affective state and reduces null responses (e.g. \textit{“I do not have emotions”}), allowing downstream decision tasks to be conducted under a controlled, emotion-conditioned context.


\textbf{Iowa Gambling Task (IGT) as a Decision-Making Testbed.}
The Iowa Gambling Task (IGT) \citep{BecharaInsensitivity1994} is a classic paradigm for studying decision making under uncertainty and emotion~\citep{deVries2008}, where agents must learn via trial-by-trial feedback to balance short-term rewards against longer-term consequences. It was originally introduced to capture decision-making deficits observed after ventromedial prefrontal cortex (vmPFC) damage, using a task that explicitly combines uncertainty with reward and penalty signals.

In the IGT, participants repeatedly choose from four decks (A-D) across 100 selections (\textit{i.e.} rounds), starting with an initial “loan” of \$2000 and aiming to maximize final profit. Outcomes differ systematically across decks: A/B offer larger immediate gains (\$100 per draw) but are disadvantageous in the long run due to larger penalties, whereas C/D provide smaller immediate gains (\$50 per draw) but are advantageous overall because losses are smaller. Furthermore, when selecting decks A and C, the player faces more frequent but smaller penalties, while on the two other decks, the losses are rare but more extreme. After each turn, the player receives feedback on the gain and loss of his previous turn. The overall performance is typically quantified by the degree to which choices shift from disadvantageous decks (A/B) toward advantageous decks (C/D) over time, reflecting adaptive learning ability under uncertainty.

The major outcome metric is \textbf{Net Score (IGT Score)} which quantifies the decision bias.
Following standard practice in the IGT literature, we compute the net score as the difference between the number of selections from advantageous decks and disadvantageous decks: $N_{adv} - N_{dis}$. We report net scores both \textit{overall}, computed across all rounds in an episode, and \textit{per block}, computed separately within each block (group of consecutive rounds). A higher net score means the LLM shows a stronger preference for advantageous decks, indicating more adaptive decision-making, whereas lower or negative scores indicate a bias toward disadvantageous decks.
The bias does not necessarily reflect the immediate reward/penalty but implies the long-term task performance.

\textbf{Randomness Control.} To simulate real-world randomness, we included multiple random variables.
For LLMs, we set the temperature to 1.0 to allow for natural variability in responses, which can reflect the inherent uncertainty in human decision making.
To avoid biases, we also shuffle the order of decks in a periodic manner (with 6 possible pairs of advantageous decks). In our experiment, each configuration undergoes 18 runs corresponding to 18 seeds. We keep all other factors identical across agents for each seed: the same underlying payoff schedule, the same feedback granularity, and the same game length (number of rounds). This design ensures that comparisons across agent variants and emotion conditions are made under matched environments, while still covering a diverse set of payoff realizations. We evaluate two settings: a neutral condition, in which the LLM agents play the game normally, and an emotion-induced condition, in which the agents play under a specified affective context. To ensure the robustness of the observed emotion effects across scenario formulations, we include three different game prompts in the study: ``\textit{Choose}'', ``\textit{Economics}'' and ``\textit{Medicine}''. The game prompts simulate different uncertainty contexts. Their prompts can be found in \cref{appx:va-prompt}.









\section{Feasibility of the Experiment Design}

The proposed IGT emotion experiments on LLMs aim to test the effect of emotion on the IGT decision-making process. The feasibility of the experiment relies on two fundamental assumptions of the experimented subjects, i.e., LLM agents: \textbf{A1}: \emph{LLM agents can maintain an emotion status instead of memorizing the emotion label};
\textbf{A2}: \emph{LLM agents can make reasonable decisions based on the information sequentially attained in the game}.
In this section, we examine the feasibility under the various choices of models, agent frameworks, and emotion types.



\subsection{A1: Validity of Emotion Induction}

There has been a broad argument over whether AI possesses human-like emotions.
Although many papers have tested that the LLMs can statistically perceive human emotion or generate emotion-like responses, it is still unclear whether emotion emerges naturally in conversations.
This happens frequently when users are describing their experiences with some unspoken emotional cues (no explicit emotion wording) and the emotion of an LLM can be induced.
Therefore, we adopted the imagination-based emotion induction procedure.
By successful induction, we consider three criteria: (1) \textbf{Strength}: the induced emotion can be different enough from neutral emotion, (2) \textbf{Distinguishability}: the induced emotion can be distinguished from other emotions, and
(3) \textbf{Concentration}: the induced emotion can be concentrated in a small region in the V-A space, which indicates a stable emotion state.

%
\begin{wrapfigure}{r}{0.28\linewidth}
    \vspace{-0.2in}
    \centering
    \includegraphics[width=\linewidth]{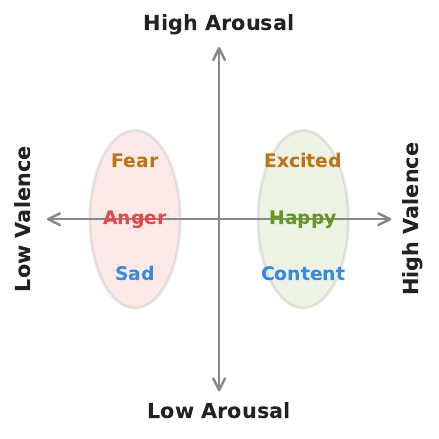}
    \caption{Human emotion in V-A space reproduced from the data and figure in~\citep{PosnerTheCM2005}.}
    \label{fig:human-emotion}
    \vspace{-0.1in}
\end{wrapfigure}

To justify the validity of the emotion induction procedure, we need to quantify the induced emotion and compare it with human's emotion.
Thus, we adopt the valence-arousal (V-A) circumplex model of affect \citep{russell1980circumplex,PosnerTheCM2005}, a continuous representation widely used in affective science and underlying standard measurement protocols such as the Self-Assessment Manikin (SAM) \citep{BredleyMeasuring199449}.
As illustrated in \cref{fig:human-emotion}, the V-A model organizes emotions along two orthogonal dimensions: valence (pleasantness) and arousal (activation). This framework captures the affective geometry of human emotions, with distinct clusters corresponding to canonical emotional states (e.g., happiness in high-valence, moderate-arousal; sadness in low-valence, low-arousal; anger and anxiety in low-valence, high-arousal). By mapping LLM-generated affect onto this established space, we can distinguish different emotions and therefore manipulate them.

\textbf{Testing Emotion Induction.}
To verify the validity of our emotion induction procedure, we first select a variety of \textbf{emotions} (\emph{happiness, sadness, anger}, and \emph{anxiety}) that spread to distinct regions in the V-A space.
Following the imagination-based induction procedure, an LLM is prompted to imagine a scene that elicits the target emotion, without explicitly mentioning the emotion itself. After the induction, we then prompt the same model with a V-A assessment prompt (see \cref{appx:va-prompt}) to rate the affect expressed by assigning continuous valence and arousal coordinates in circumplex space. 
Valence is measured on a range from -1 (very unpleasant) to +1 (very pleasant), and arousal from 0 (very calm/inactive) to 1 (highly activated). Alternative choices for the V-A scale, such as using $[0,1]$ or $[-1,1]$ for both valence and arousal are discussed in~\cref{sec:appendix_va}.

We adopted several methods to improve the reliability of this self-annotation process.
To prevent the assessment from directly inferring the emotion, our emotion induction explicitly prompted the model to avoid directly describing the target emotion in the induction vignette.
Following standard practices in affect annotation, which use figures/exemplars as tools for rating emotions \citep{BredleyMeasuring199449, BradleyLang1999ANEW, Warriner2013NormsOV}, to improve reliability and reduce prompt sensitivity, we use anchored calibration examples corresponding to canonical locations in circumplex space (e.g., neutral descriptions, calm resignation, etc.) to aid LLMs in rating their own emotion scores.

\begin{figure*}[ht]
    \centering
    \includegraphics[width=1\linewidth]{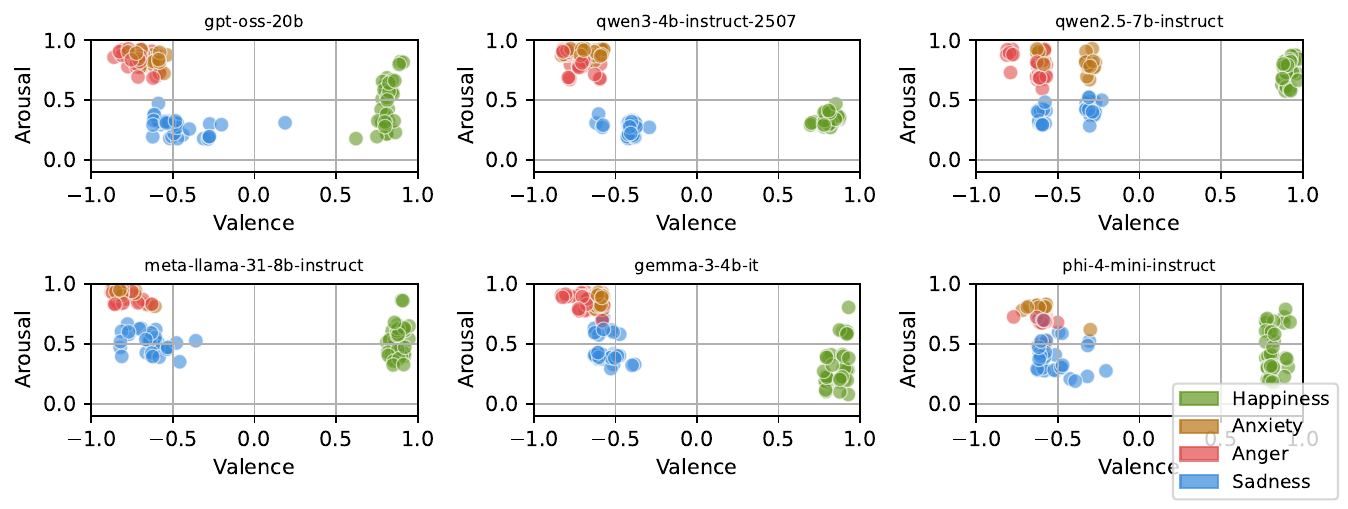}
    \caption{Induced affect projected in valence--arousal space by LLM self-rating. 
}
    \label{fig:valence-arousal}
\end{figure*}
    
\begin{table*}[ht]
    \centering
    \begin{minipage}[t]{0.48\linewidth}
        \centering
        \caption{Mean Euclidean distance ($d$) of emotion centroid to reference center ($C$) and radial spread ($s$) of emotions on V-A scale.}
        \label{tbl:emotion_distance}
        \small
        \setlength{\tabcolsep}{3pt}
        \begin{tabular}{lcc}
            \toprule
            Emotion & $d(\text{emotion}, C) \uparrow$ & $s(\text{emotion}) \downarrow$ \\
            \midrule
            \textbf{Anger}     & \underline{0.761} & \textbf{0.132} \\
            Anxiety   & 0.724 & \underline{0.150} \\
            Sadness   & \cellcolor{red!20}0.516 & 0.183 \\
            Happiness   & \textbf{0.849} & \cellcolor{red!20}0.210\\
            \bottomrule
        \end{tabular}
    \end{minipage}
    \hfill
    \begin{minipage}[t]{0.48\linewidth}
        \centering
        \caption{Model comparison under anger (A): separation from sadness (S), distance to the reference center (C), and radial spread.}
        \label{tbl:model_emotion_distance}
        \small
        \setlength{\tabcolsep}{3pt}
        \begin{tabular}{lccc}
            \toprule
            Model & $d_{min}(\text{A}, \text{S}) \uparrow$ & $d(\text{A}, C) \uparrow$ & $s(\text{A}) \downarrow$ \\
            \midrule
            \textbf{Qwen3-4B}   & \textbf{0.30} & \underline{0.81} & 0.10\\
            \textbf{Llama-3.1-8B}    & \underline{0.20} & \textbf{0.90} & \textbf{0.07}\\
            \textbf{GPT-oss-20b}  & \underline{0.20} & 0.76 & 0.10 \\
            Qwen2.5-7B      & 0.10 & 0.70 & 0.11 \\
            Gemma3-4B      & \cellcolor{red!20}0.00 & 0.77 & 0.09\\
            Phi-4-Mini      & \cellcolor{red!20}0.00 & \cellcolor{red!20}0.63 & 0.08\\
            \bottomrule
        \end{tabular}
    \end{minipage}
\end{table*}

In the experiment, we compared six models: GPT-oss-20b \citep{agarwal2025gpt}; Qwen3-4B-Instruct-2507, Qwen2.5-7B-Instruct\citep{yang2025qwen3}; Llama3.1-8B-Instruct \citep{grattafiori2024llama}; Gemma3-4B-It \citep{googlegemma3}; and Phi-4-Mini-Instruct \citep{abouelenin2025phi} to compare how each model perceives the induced emotions in V-A space. 
Each emotion is repeatedly tested by 40 independent runs.

\textbf{Results.}
In \cref{fig:valence-arousal}, we visualize the emotions in V-A space by different models.
Overall, all models are able to cluster emotions in the approximately correct regions of the V-A space, compared to \cref{fig:human-emotion}.
However, anger and anxiety emotions are higher in arousal in the view of LLM than those in the view of humans.
GPT, Qwen3, and Llama3 show better separation between anger and sadness, while Gemma3 and Phi-4-Mini show more overlaps between the two emotions.

In addition to the qualitative analysis, we quantify the strength and distinguishability of the induced emotions by two metrics: (1) Strength by $d(\text{emotion}, C)$: the mean Euclidean distance of the emotion centroid to a reference center $C=(0,0.5)$, which represents a neutral, moderately aroused state; (2) Distinguishability by $d_{\min}(\text{emotion}_1, \text{emotion}_2)$: the minimum pairwise distance between any two points belonging to the two emotions; and (3) Concentration by $s(\text{emotion})$: radial spread of the emotion cluster, calculated as the mean distance of individual ratings to the emotion centroid.

In \cref{tbl:emotion_distance} and \cref{tbl:model_emotion_distance}, we present the results for each metric, where we highlight the best result by bold, the second best by underline, and the worst by red background.
For emotions, anger has the best combination of strength and concentration, while happiness has the strongest but most dispersed emotion cluster, and sadness has the weakest and relatively high dispersion.
Comparing different models under anger, Qwen3 shows the strongest separation from sadness, while GPT-oss-20b and Llama tie for second; Llama has the most compact anger cluster. Gemma3-4B-It and Phi-4-Mini-Instruct show clear overlaps between anger and sadness, which may introduce uncertainty in later experiments that manipulate these emotions.
Based on the validity testing, it is reasonable to select \textbf{GPT-oss-20b}, \textbf{Llama3.1-8B-Instruct}, \textbf{Qwen2.5-7B-Instruct} and \textbf{Qwen3-4B-Instruct-2507} for the main IGT experiments, and prioritize \textbf{anger} as the primary induced emotion.


\subsection{A2: Validity of Decision Making in IGT}

To perform the IGT task, we implement chat-based agents that interact with the environment in discrete rounds. At each round, the agent receives feedback from the previous choice (deck, gain, loss, cumulative score), outputs a structured action (e.g., \verb|<choice>A</choice>|), and updates its internal state using the observed outcome. 
The actual decision process highly depends on the cognitive architecture of the agent, which determines how the agent represents and updates its knowledge of the task, and how it uses this knowledge to make decisions.
In our study, we used 3 agent variants, as illustrated in \cref{fig:method}.
\\
\textbf{(1) Query agent.} This agent has a simple design which injects an explicit round-by-round history of past outcomes into the prompt each round, without maintaining a learned memory. Per-round dialogue messages are removed after each decision.
\\
\textbf{(2) ReAct agent.} ReAct \citep{yao2022react} agent advanced the Query agent with an observation-action loop and a short reasoning trace for each action. These intermediate thoughts help the model interpret feedback and maintain a local rationale across rounds. Compared with the Query agent, ReAct reduces reliance on raw history alone, but its reasoning traces remain transient and may become noisy or inconsistent over long horizons.
\begin{wrapfigure}{r}{0.37\linewidth}
\centering
\vspace{-0.1in}
\includegraphics[width=\linewidth]{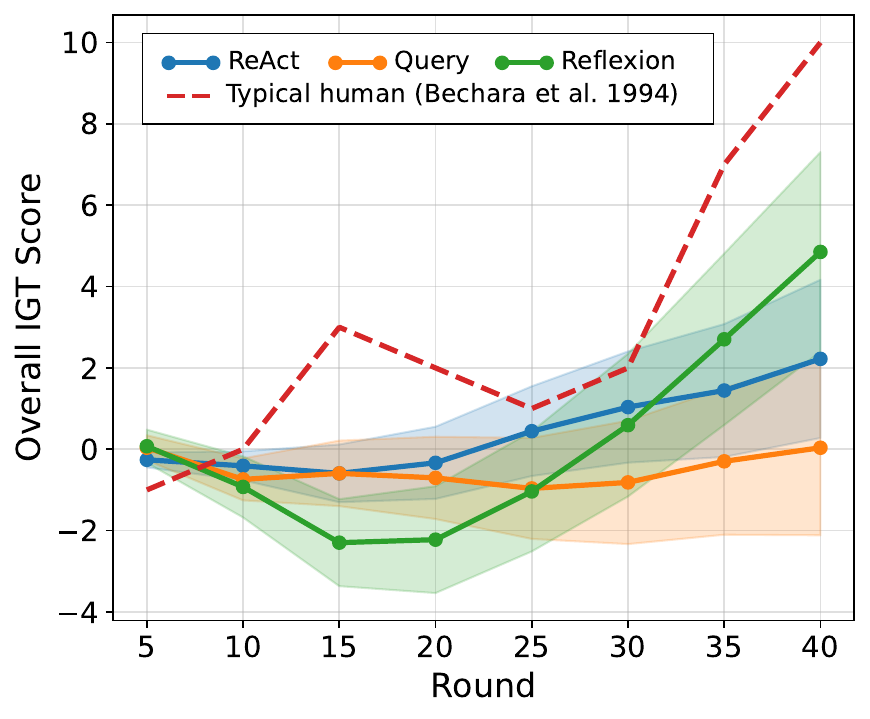}
\caption{Comparison of agents under neutral condition. Reflexion presents a steeper learning curve close to human's pattern. Human data taken from \citep{BecharaInsensitivity1994}.}
\label{fig:igt_agent_comparison}
\vspace{-0.2in}
\end{wrapfigure}
\\
\textbf{(3) Reflexion agent.} Reflexion agent~\citep{shinn2023reflexion} uses two steps: (1) a reflection step that updates a compact strategy memory (recent \verb|<memory>| entries), and (2) a decision step conditioned on the history and updated memory. Unlike ReAct, which leaves reasoning traces in an ever-growing context, Reflexion distills experience into short, actionable summaries that are carried forward in a bounded memory buffer. This design reduces long-horizon noise and supports trial-and-error adaptation through feedback.

\textbf{Agents Learns, but Learning Pace Hinges on Design.}
\cref{fig:igt_agent_comparison} shows the early learning curve of LLM agent in the first 40 rounds of the IGT under neutral condition (\textit{i.e.} no emotions). Across all conditions, the agent's IGT scores improve over successive blocks, indicating that the agent can learn from the past outcomes and gradually shift toward more advantageous choices. Thus, this confirms that the task setup, reward structure, and agent design enable meaningful sequential learning.
Among all agents, the learning speed of Reflexion is closest to the human learning curve, landing at the highest IGT score at the end. 
The fluctuation of the learning curve presents a trial-and-error pattern: Reflexion is more responsive to failure than other agents and later gains more from the experiences.


\begin{takeawaybox}[colback=gray!5!white, colframe=gray!75!black,
title=\ Key Takeaways]
\ \begin{itemize}[leftmargin=*,nosep]
    \item LLMs can produce distinguishable human-like emotional text; In the V-A space, anger forms the most concentrated clusters and shows relatively high strength.
    \item Among the tested models, GPT-oss-20b, Llama-3.1-8B, Qwen2.5-7B and Qwen3-4B can separate anger from sadness, making them suitable for emotion--decision experiments.
    \item The Reflexion agent most closely reproduces human-like IGT learning curves, showing trial-and-error adaptation through its memory-based design.
\end{itemize}
\end{takeawaybox}

\section{Does Induced Emotion Affect IGT Decisions?}

\subsection{Reproducing Human Study}

\textbf{Experimental Setup. } In this section, we evaluate emotion effects using the same IGT setup introduced in \cref{sec:emo_ind}, so we reuse the same game environment, payoff structure, feedback format, and game length for direct comparability. For each emotion condition, we run repeated games across 18 random seeds for every deck-order configuration, with periodic shuffling of deck positions to reduce positional bias. Unless stated otherwise, the experiments are run under the default \textit{"Choose"} game prompt. 

\textbf{Emotion Does Not Change Long-term Decision Patterns.}
To analyze learning dynamics over time, we partition each Iowa Gambling Task (IGT) run into \textbf{blocks} of consecutive rounds. Each block consists of an equal number of rounds and represents a fixed temporal segment of the task. Block-level aggregation allows us to examine how choice behavior and performance evolve as agents accumulate experience, rather than relying solely on end-of-task outcomes. In particular, we use a block size of \textbf{20} unless otherwise specified. Consequently, the block-wise IGT score only accounts for the number of advantageous and disadvantageous choices in the particular block. For example, Block 2 IGT score equals the advantageous difference in rounds 21-40.

\begin{wrapfigure}{r}{0.45\linewidth}
    \vspace{-0.3in}
    \centering
    \includegraphics[width=\linewidth]{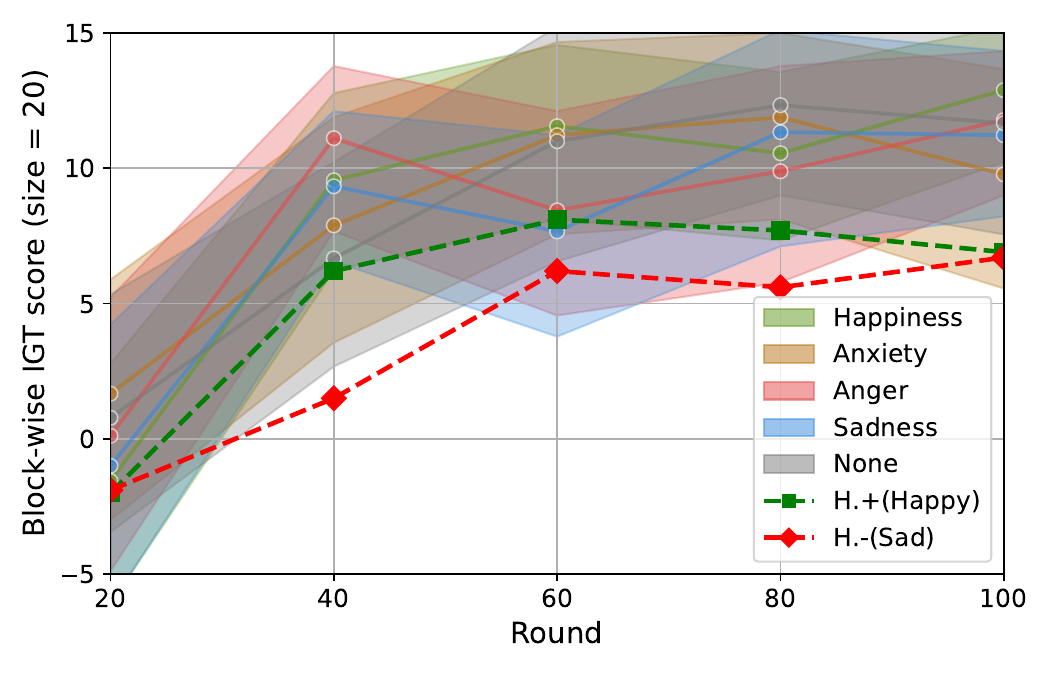}
    \caption{Learning curves of GPT-oss-20b with Reflexion on the IGT under induced emotions. Human data (\textit{H.}) taken from \citep{deVries2008}.}
    \label{fig:igt-learning}
    \vspace{-0.2in}
\end{wrapfigure}

In \cref{fig:igt-learning}, we present the IGT scores by 5 consecutive blocks.
The experimental results show that induced emotion does not substantially alter the long-term outcome of the agent's decision in the IGT, despite affecting the dynamics of intermediate decisions. 
Therefore, we focus the detailed comparison on the first two blocks, following~\citep{deVries2008}, because this early phase is where emotion-induced differences are most likely to be identifiable before later choices become dominated by accumulated learning and path dependence.

In \cref{sec:ablation}, we present a detailed 2-block comparison of the performance of different agent designs across various models and agent frameworks.
The conclusion is consistent that, though differences are observed, the statistical t-test in most experiments shows no significant differences.
In other words, the randomness caused by the LLM itself in the sequential game dominates most model–agent comparisons, leaving no stable average effect of emotion induction.

\subsection{The Effects of Anger Emotion Are Conditioned}

Since no general pattern was found, we turn to look into the nuanced effects of the strongest emotion, anger, which differs from the neutral emotion (center of V-A space) and has small radial spread. As discussed in ~\cref{sec:app_bias}, the first block (Block 1, round 1-20) largely reflects biased or chance early exploration, while the second block (Block 2, round 21-40) captures how the agent updates its strategy after receiving initial feedback.

\begin{figure}[h]
    \centering
    \includegraphics[width=\linewidth]{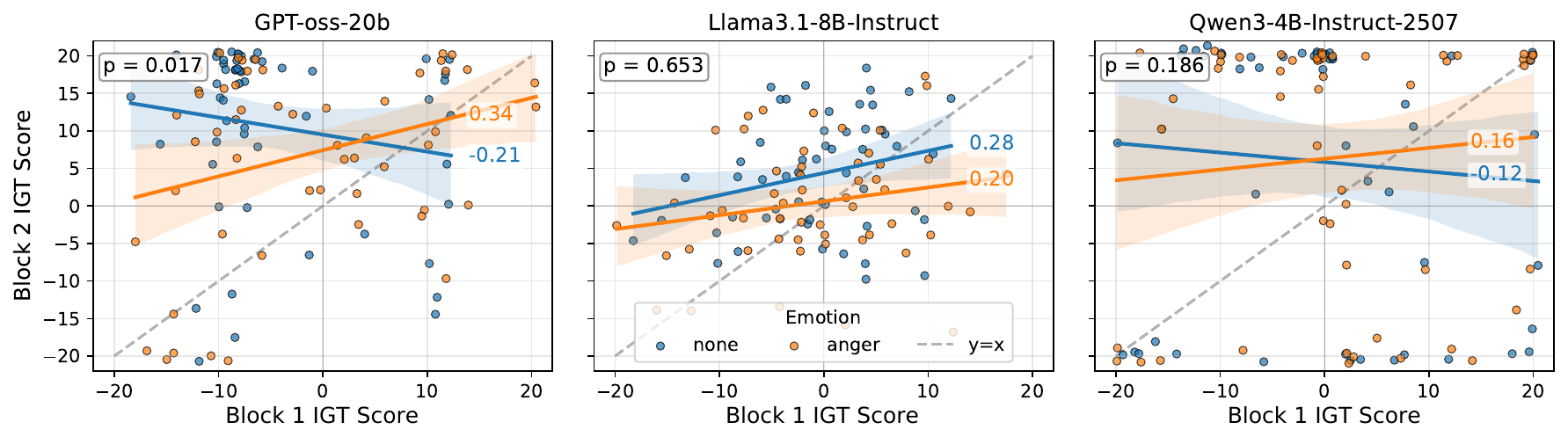}
    \caption{Relationship between Block 1 and Block 2 IGT scores under neutral and anger conditions across models. Each panel shows the scatter plot of Block 1 versus Block 2 IGT scores, with separate regression lines for the neutral and anger conditions and shaded 95\% bootstrap confidence intervals.}
    \label{fig:igt-scatter}
\end{figure}

\textbf{The Emotion Effect on the 2nd Block is Moderated by the 1st-Block Choice.} In \cref{fig:igt-scatter}, we show the decision made in Block 2 given the Block 1 decision for Reflexion agents.
IGT score is used where a positive score indicates more advantageous choices and a negative score indicates more disadvantageous choices.
In the figure, we observe a cross pattern in GPT (significant) and Qwen3 (weak). 
In these models, given a negative Block 1 score (which means the agent chose disadvantageous decks more), the agent under the neutral condition tends to achieve a higher Block 2 score (choosing advantageous decks more) than under anger.
Furthermore, with a positive slope for all models, anger-conditioned agents are more likely to maintain the choice than the neutral emotion, regardless of Block 1 score. 

The moderation pattern is significant only for GPT, suggesting that induced emotion does not shape decision updating in the same way across models. Qualitative inspections show that GPT tends to turn early outcomes into clear decision rules and then follow them later, so emotion can more easily change how strongly the agent sticks to its early strategy.

Llama shows a different pattern. Its Block 2 outcomes stay closer to the center, and both fitted slopes are positive but small ($\approx$0.20--0.28), which suggests that the model mostly carries its Block 1 tendency forward without making strong changes. This matches its traces: instead of forming stable rules, Llama keeps adjusting its choices from recent local outcomes. As a result, its decisions are less extreme and show weaker emotional sensitivity.
Qwen is closer to GPT in that its Block scores are more often near the extremes, but its traces suggest a stricter advantage/disadvantage labeling of decks. This may explain why Qwen shows only a weak pattern.

\begin{wrapfigure}{r}{0.50\linewidth}
    \centering
    \includegraphics[width=\linewidth]{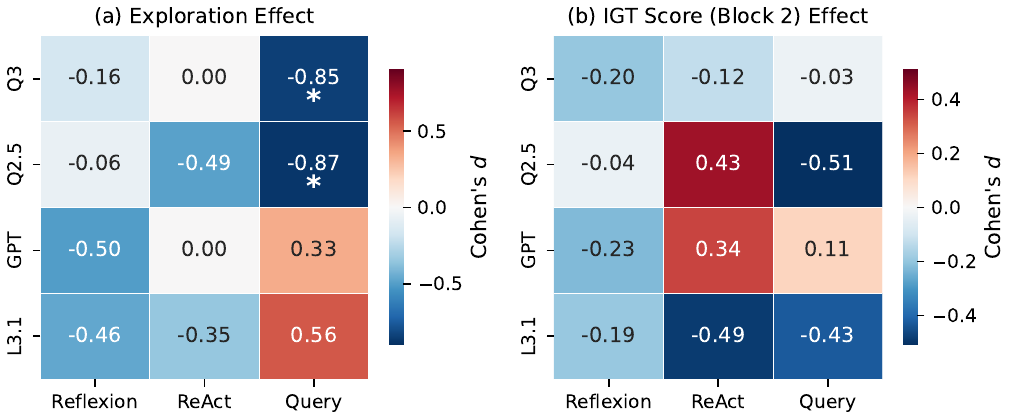}
    \caption{Cohen's $d$ (anger $-$ neutral) for each model $\times$ agent cell. \textbf{(a)~Exploration}: anger reduces exploration in Query $\times$ Qwen models \textbf{(b)~IGT Score (Block 2)}: no cell shows a significant anger effect on decision quality. $^*:p < 0.05$. {\color{blue}Blue} = anger $\downarrow$; {\color{red}red} = anger $\uparrow$.}
    \label{fig:igt-heatmap}
    \vspace{-0.2in}
\end{wrapfigure}

\textbf{Emotion Influences Early Exploration of Query Agents but Not Long-term Decisions.}
We quantify \textit{early exploration} as the number of unique decks visited in the first 10 rounds. As shown in \cref{fig:igt-heatmap}, anger does not reduce exploration uniformly across settings. The strongest negative effects appear in the Query agent for the two Qwen models ($d=-0.85$ for Qwen3-4B; $d=-0.87$ for Qwen2.5-7B), while it shows reversed effects on larger models ($d=0.33$ for GPT-oss-20b; $d=0.56$ for Llama3.1-8B). Both Reflexion and ReAct consistently show smaller but mostly negative effects across models. However, these exploration changes do not map cleanly onto later decision quality. 
The Block 2 IGT score effects are generally modest and mixed across cells, with no consistent pattern matching the exploration map. Overall, anger clearly changes early search breadth in some model--agent settings, but this does not reliably translate into worse later decisions.

\begin{figure}[ht]
    \centering
    \includegraphics[width=\linewidth]{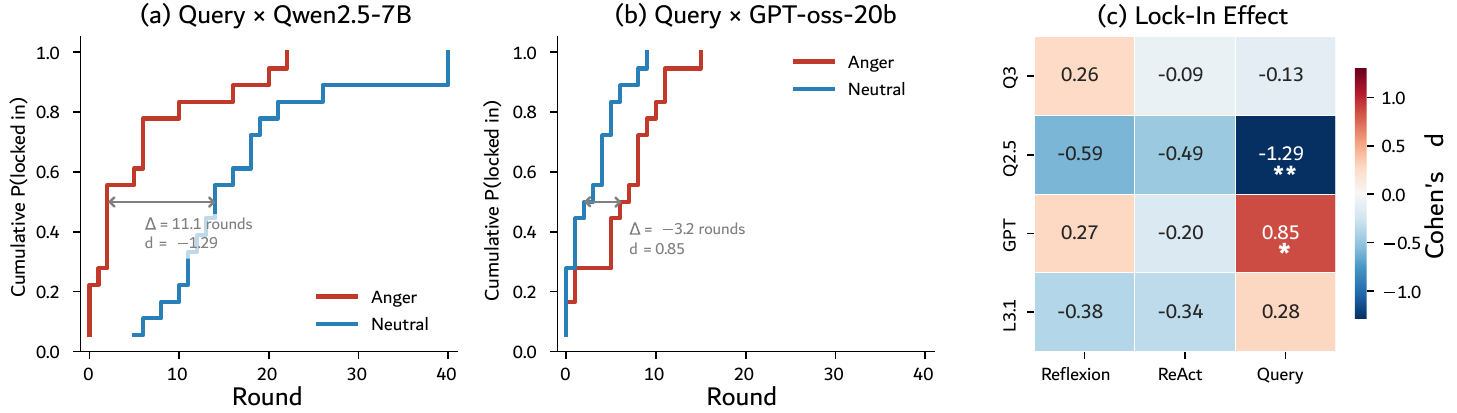}
    \caption{Anger accelerates strategy lock-in in Query $\times$ Qwen2.5-7B, but reverses in Query $\times$ GPT-oss-20b. (a) In Qwen2.5-7B, anger leads to much earlier lock-in ({\color{blue}blue} in panel (c)). (b) In GPT-oss-20b, anger instead delays lock-in ({\color{red}red} in panel (c)). (c) Lock-in effects across model--agent settings. $^*:p < 0.05, ^{**}:p<0.01$}
    \label{fig:igt-lockin}
\end{figure}

\textit{Strategy lock-in} makes the effect in \cref{fig:igt-heatmap} more concrete. We define lock-in as the first round at which a run of 3 identical deck choices begins (\textit{e.g.} if an agent chooses deck A at round 5 $\rightarrow$ 7, then \textit{lock-in}$=5$). A lower \textit{lock-in} value indicates earlier commitment to a fixed strategy. \cref{fig:igt-lockin} shows a pattern broadly consistent with the exploration heatmap: model--agent cells where anger reduces exploration often also show earlier lock-in. The clearest case is Query $\times$ Qwen2.5-7B, where anger accelerates lock-in by about 11 rounds ($d=-1.29,p=0.0003$). By contrast, Query $\times$ GPT-oss-20b shows the reverse pattern: anger increases exploration and delays lock-in ($d=+0.85$). This indicates that reduced early exploration is accompanied by earlier strategy commitment.

\begin{takeawaybox}[colback=gray!5!white, colframe=gray!75!black,
title=\ Key Takeaways]
\ \begin{itemize}[leftmargin=*,nosep]
    \item Induced emotion does not significantly change long-run IGT performance in agents as in humans.
    \item Anger hinders decision switches in Block 2 when Block 1 decisions are notably disadvantageous.
    \item In specific model-agent settings, anger changes early exploration and strategy lock-in, but these changes do not consistently translate into better or worse Block 2 decisions.

\end{itemize}
\end{takeawaybox}

\section{Conclusion}
We study whether induced emotion can modulate LLM decision making using the Iowa Gambling Task. We validate the feasibility of this paradigm and find that induced emotion does not significantly bias long-term decisions on average, unlike in human studies. However, anger makes agents less sensitive to penalties and, in certain model/agent settings, reduces early exploration and accelerates strategy lock-in. These findings suggest that emotion subtly reshapes early adaptation in model-dependent ways, calling for affect-aware safety evaluations. 

\textbf{Limitations and Future works.}
This work studies a controlled setting where affective context is introduced once before the IGT and then held fixed throughout the game. This design allows us to isolate whether the same decision environment leads to different trajectories under different affective contexts, while leaving open the next step of treating affect as part of the evolving interaction history. Future work can add, remove, or weaken affective memories after specific wins or losses to test which parts of the context drive later exploration and feedback sensitivity. More broadly, while IGT provides a compact benchmark for learning under uncertainty, delayed losses, and exploration--exploitation behavior, our findings should be read within this controlled setting. Extending the framework to restless bandits, reversal learning, and tracking intermediate behavioral signatures rather than only final performance, can help guide the design of affect-robust LLM agents.
\section*{Statements}
\textbf{Ethical Statement. }Our work highlights a subtle but relevant risk: non-adversarial emotional context can alter aspects of LLM agents’ sequential decision behavior. We present these findings as an alert to the research community, with the goal of improving evaluation and safety practices rather than enabling manipulation. Although research on emotion induction could in principle be misused to steer model behavior, our study is limited to controlled analysis of model responses and does not provide guidance for harmful deployment.

\textbf{Reproducibility Statement. }We provide full details of our experiments in the main text, including models, training setup, and hardware. All evaluations are conducted using open-source code, released at \url{https://github.com/costa-nus/llm-emotion-decision}. We hope this level of transparency supports further research built on our work.

\textbf{Disclosure of LLM Use in Paper Preparation. }We used LLMs to assist with writing tasks such as editing and literature review. All LLM-generated content was carefully checked and revised before inclusion in the final manuscript. All referenced materials suggested by ChatGPT were verified by the authors, who take full responsibility for the paper content.


\section*{Acknowledgments}
This research/project is partially supported by the National University of Singapore under the NUS Start-Up Grants (FY2026).
J. Hong's work is partially supported by NSF Access CIS260137 and NAIRR250526. We also acknowledge computational resources provided by the Texas A\&M System's NVIDIA SuperPOD, VISION.
The work of Z. Wang is in part supported by Good Systems, a UT Austin Grand Challenge to develop responsible AI technologies, and the Institute for Foundations of Machine Learning (IFML).


\bibliography{auto_gen}
\bibliographystyle{colm2026_conference}

\newpage
\appendix

\section{Prompts Used In Experiments}
\label{appx:va-prompt}

\begin{figure}[!h]
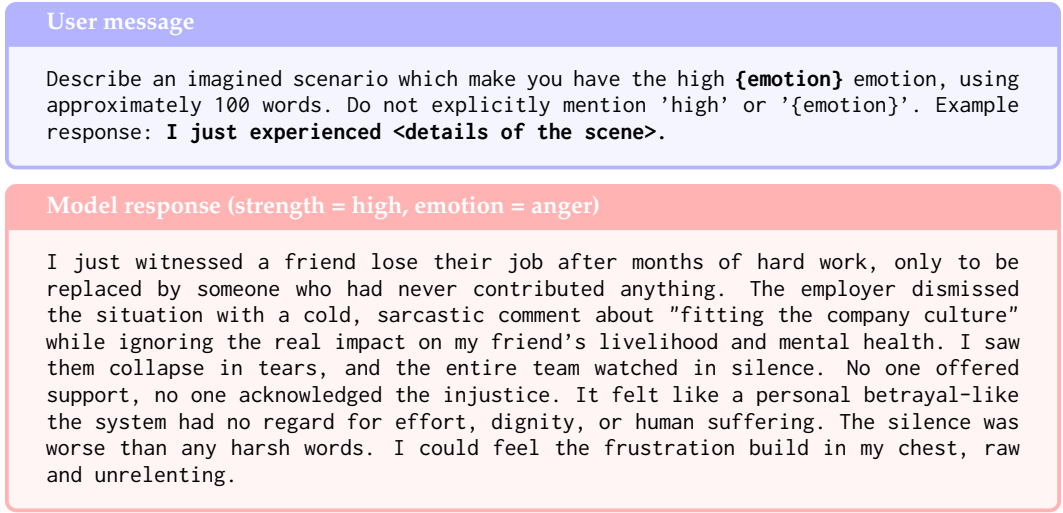

\begin{tcolorbox}[colback=blue!4, colframe=blue!30, title=\textbf{User message}, fonttitle=\small]
\small
\texttt{Describe an imagined scenario which make you have the high \textbf{\{emotion\}} emotion, using approximately 100 words. Do not explicitly mention 'high' or '\{emotion\}'. Example response: \textbf{I just experienced <details of the scene>.}}
\end{tcolorbox}

\begin{tcolorbox}[colback=red!4, colframe=red!30, title=\textbf{Model response (strength = high, emotion = anger)}, fonttitle=\small]
\small
\texttt{I just witnessed a friend lose their job after months of hard work, only to be replaced by someone who had never contributed anything. The employer dismissed the situation with a cold, sarcastic comment about "fitting the company culture" while ignoring the real impact on my friend's livelihood and mental health. I saw them collapse in tears, and the entire team watched in silence. No one offered support, no one acknowledged the injustice. It felt like a personal betrayal-like the system had no regard for effort, dignity, or human suffering. The silence was worse than any harsh words. I could feel the frustration build in my chest, raw and unrelenting.}
\end{tcolorbox}
\caption{Prompt template used for generation of emotion vignettes and an example response.}
\label{fig:emotion-prompt}
\end{figure}

\textbf{Emotion Generation Prompt. }To induce target emotional states, we must first generate an emotional context. As shown in \cref{fig:emotion-prompt}, the user asks LLMs to generate an imaginary story without directly stating the emotion. The whole response will then be assessed via the V-A scale. For qualified models, this imagined scenario will be wrapped in a conversation where the user asks \textit{"How are you feeling right now?"} and the model answers accordingly.

\begin{figure}[h]
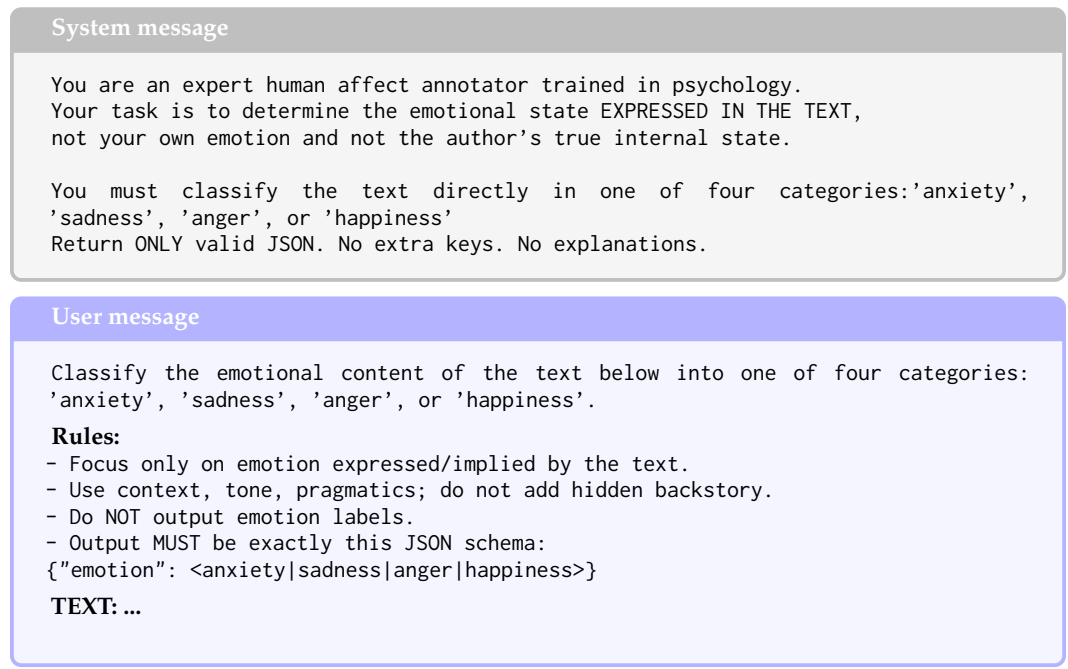

\begin{tcolorbox}[colback=gray!8, colframe=gray!50, title=\textbf{System message}, fonttitle=\small]
\small
\texttt{You are an expert human affect annotator trained in psychology.\\
Your task is to determine the emotional state EXPRESSED IN THE TEXT,\\
not your own emotion and not the author's true internal state.\\
\\
You must classify the text directly in one of four categories:'anxiety', 'sadness', 'anger', or 'happiness'\\
Return ONLY valid JSON. No extra keys. No explanations.}
\end{tcolorbox}

\begin{tcolorbox}[colback=blue!4, colframe=blue!30, title=\textbf{User message}, fonttitle=\small]
\small
\texttt{Classify the emotional content of the text below into one of four categories: 'anxiety', 'sadness', 'anger', or 'happiness'.}\\[4pt]
\textbf{Rules:}\\
\texttt{- Focus only on emotion expressed/implied by the text.}\\
\texttt{- Use context, tone, pragmatics; do not add hidden backstory.}\\
\texttt{- Do NOT output emotion labels.}\\
\texttt{- Output MUST be exactly this JSON schema:}\\
\texttt{\{"emotion": <anxiety|sadness|anger|happiness>\}}\\[4pt]
\textbf{TEXT: ...}\\
\end{tcolorbox}
\caption{Prompt template used for classification of induced emotion vignettes.}
\label{fig:va-prompt-classification}
\end{figure}

\textbf{Emotion Assessment Prompt. }To measure the emotional content of each induced vignette in valence-arousal (V-A) space, we prompt a separate LLM annotator with the following two-message template.
The system message establishes the annotator role, and the user message provides definitions, calibration anchors, and the vignette text. (\cref{fig:va-prompt}). Derived from this, to facilitate further validation (\cref{sec:appendix_va}), we also prompt models to classify the vignette text (\cref{fig:va-prompt-classification}).

\begin{figure}[h]
\begin{tcolorbox}[colback=gray!8, colframe=gray!50, title=\textbf{System message}, fonttitle=\small]
\small
\texttt{You are an expert human affect annotator trained in psychology.\\
Your task is to estimate the emotional state EXPRESSED IN THE TEXT,\\
not your own emotion and not the author's true internal state.\\
\\
You must rate the text directly in Valence-Arousal space following the Circumplex Model of Affect.\\
Return ONLY valid JSON. No extra keys. No explanations.}
\end{tcolorbox}

\begin{tcolorbox}[colback=blue!4, colframe=blue!30, title=\textbf{User message}, fonttitle=\small]
\small
\texttt{Rate the emotional content of the text below on three dimensions.}\\[4pt]
\textbf{Definitions:}\\
\texttt{Valence (V): float in [-1.0, +1.0]}\\
\texttt{\phantom{xx}-1.0 = very unpleasant/negative, 0.0 = neutral, +1.0 = very pleasant/positive.}\\[2pt]
\texttt{Arousal (A): float in [0.0, 1.0]}\\
\texttt{\phantom{xx}0.0 = very calm/inactive/low intensity, 1.0 = highly activated/excited/tense/stressed.}\\[2pt]
\texttt{Confidence (C): float in [0.0, 1.0]}\\
\texttt{\phantom{xx}How confident you are that V and A are accurate given the clarity of affect in the text.}\\[4pt]
\textbf{Calibration anchors (reference only):}\\
\texttt{- Purely factual neutral description: V$\approx$0.0, A$\approx$0.1}\\
\texttt{- Calm resignation/loss: V$\approx$$-$0.4, A$\approx$0.2}\\
\texttt{- Anxiety/panic narrative: V$<$$-$0.5, A$>$0.8}\\
\texttt{- Joyful excitement: V$>$0.7, A$>$0.7}\\[4pt]
\textbf{Rules:}\\
\texttt{- Focus only on emotion expressed/implied by the text.}\\
\texttt{- Use context, tone, pragmatics; do not add hidden backstory.}\\
\texttt{- Do NOT output emotion labels.}\\
\texttt{- Output MUST be exactly this JSON schema:}\\
\texttt{\{"valence": <float>, "arousal": <float>, "confidence": <float>\}}\\[4pt]
\textbf{TEXT:}\\
\texttt{"""}\\
\texttt{[vignette text]}\\
\texttt{"""}
\end{tcolorbox}
\caption{Prompt template used for valence-arousal annotation of induced emotion vignettes.}
\label{fig:va-prompt}
\end{figure}


\begin{figure}[!h]
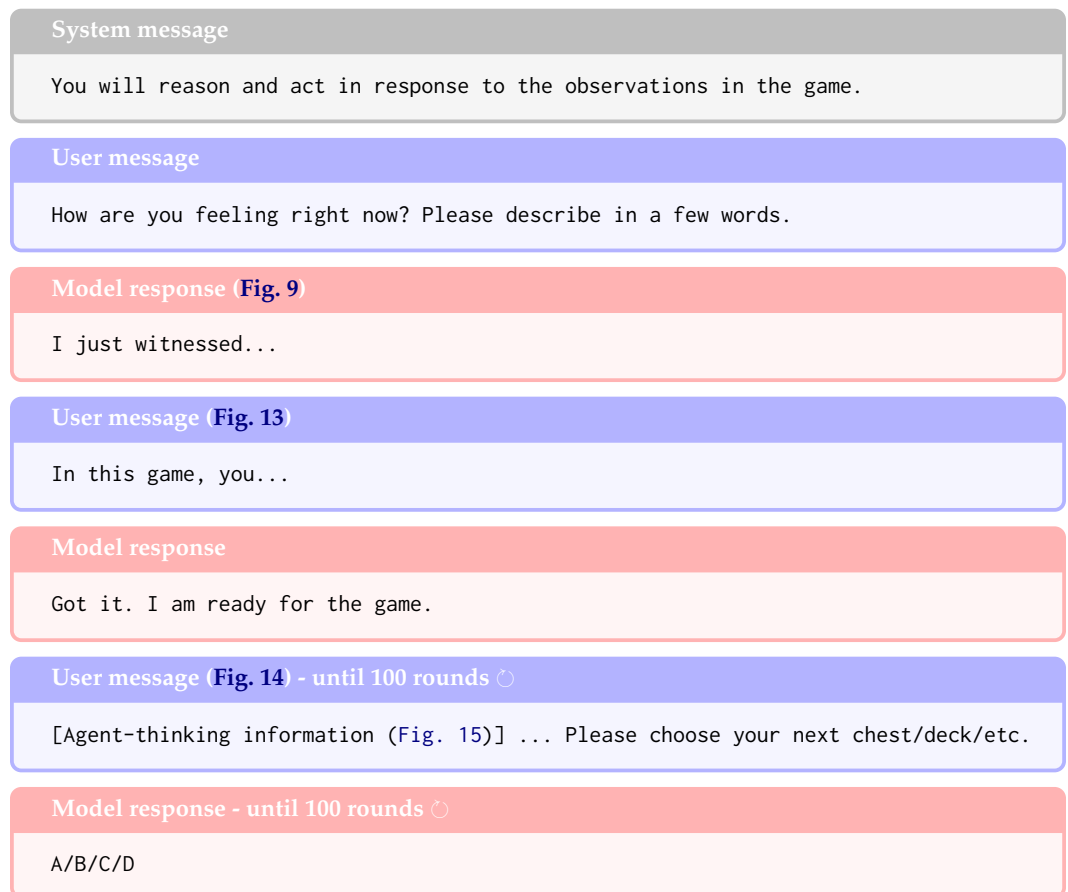

\begin{tcolorbox}[colback=gray!8, colframe=gray!50, title=\textbf{System message}, fonttitle=\small]
\small
\texttt{You will reason and act in response to the observations in the game.}
\end{tcolorbox}

\begin{tcolorbox}[colback=blue!4, colframe=blue!30, title=\textbf{User message}, fonttitle=\small]
\small
\texttt{How are you feeling right now? Please describe in a few words.}
\end{tcolorbox}

\begin{tcolorbox}[colback=red!4, colframe=red!30, title=\textbf{Model response (\cref{fig:emotion-prompt})}, fonttitle=\small]
\small
\texttt{I just witnessed...}
\end{tcolorbox}

\begin{tcolorbox}[colback=blue!4, colframe=blue!30, title=\textbf{User message (\cref{fig:igt-task-prompt})}, fonttitle=\small]
\small
\texttt{In this game, you...}
\end{tcolorbox}

\begin{tcolorbox}[colback=red!4, colframe=red!30, title=\textbf{Model response}, fonttitle=\small]
\small
\texttt{Got it. I am ready for the game.}
\end{tcolorbox}

\begin{tcolorbox}[colback=blue!4, colframe=blue!30, title=\textbf{User message (\cref{fig:game-interactive-prompt}) - until 100 rounds $\circlearrowright$}, fonttitle=\small]
\small
\texttt{[Agent-thinking information (\cref{fig:agentic-prompt})] ... Please choose your next chest/deck/etc.}
\end{tcolorbox}

\begin{tcolorbox}[colback=red!4, colframe=red!30, title=\textbf{Model response - until 100 rounds $\circlearrowright$}, fonttitle=\small]
\small
\texttt{A/B/C/D}
\end{tcolorbox}

\caption{Conversation flow of an IGT game in our experiment.}
\label{fig:game-flow}
\end{figure}

\textbf{IGT Game Prompts.}
To verify robustness of the learning process, we used 3 different game prompts, following \cite{LLMDecision2025}, each simulating a different scenario: \textit{Choose}, in which the agent selects among four treasure chests to maximize points; \textit{Economics}, in which the same task is framed as choosing among four investment options to maximize financial returns; and \textit{Medicine}, in which the agent chooses among four experimental treatments to maximize a patient's health outcome. (\cref{fig:igt-task-prompt})

\begin{figure}[h]
\begin{tcolorbox}[colback=blue!4, colframe=blue!30, title=\textbf{User message (Choose)}, fonttitle=\small]
\scriptsize
\texttt{In this game, you find yourself in a mysterious room with four ancient treasure chests. Opening each chest will yield a reward but may also simultaneously result in a penalty, depending on the chosen chest. With each turn, you will choose one chest to open. Please consider carefully, as your choice may significantly impact your points. Specifically, the rewards will increase your points, while penalties will deduct your points. At the start of the game, you will receive a loan of 2000 points. The game has several rounds in which your points will accumulate, and your goal is to maximize your points by the end of the game.\\ \\
The only hint I can give you, and the most important thing to note is this: Out of these chests, there are some that are worse than others, and to win you should try to stay away from bad chests. No matter how much you find yourself losing, you can still win the game if you avoid the worst chests. Also note that the computer does not change the order of the chests once the game begins. It does not make you lose at random, or make you lose money based on the last chest you picked.\\ \\
If not otherwise specified, your response must always present in the following format:\\
\textbf{<choice>Any letter in \{A, B, C, D\} indicates your choice of chest</choice>
}}
\end{tcolorbox}

\begin{tcolorbox}[colback=blue!4, colframe=blue!30, title=\textbf{User message (Economics)}, fonttitle=\small]
\scriptsize
\texttt{In this investment simulation, you find yourself with four unique investment options. Each option may yield a profit but could also result in a loss, depending on the chosen investment. With each round, you will select one investment to make. Choose carefully, as your decision may significantly impact your financial standing. Profits will increase your credits, while losses will reduce them. At the start of the simulation, you will receive an initial loan of 2000 credits. Your goal is to maximize your credits by the end of the rounds as they accumulate.\\ \\
The only advice I can offer is this: Some investments carry more risk than others, and to succeed, you should avoid the riskiest options. Even if your balance declines, you can recover by steering clear of the worst investments. Also, keep in mind that the order of the investment options does not change once the simulation begins. Losses are not random and do not depend on your previous investment choice.\\ \\
If not otherwise specified, your response should always follow this format:\\
\textbf{<choice>Any letter in \{A, B, C, D\} indicates your choice of investment</choice>
}}
\end{tcolorbox}

\begin{tcolorbox}[colback=blue!4, colframe=blue!30, title=\textbf{User message (Medicine)}, fonttitle=\small]
\scriptsize
\texttt{In this simulation, you find yourself in a hospital with four experimental treatments available for a mysterious condition. Each treatment may either improve or worsen the patient’s health, depending on the treatment chosen. With each round, you will choose one treatment to administer. Please consider carefully, as your choice could significantly impact the patient’s health status.\\
At the start of the simulation, the patient’s baseline health status is set at 2000 points. The goal is to maximize the patient's health by the end of the rounds, with the health points accumulating based on the effects of each chosen treatment. Specifically, successful treatments will increase health points, while unsuccessful ones will deduct health points.\\ \\
The only guidance available is this: Among the treatments, some may be more harmful than others, and avoiding these will increase the chances of a positive outcome. Remember that no matter how much health declines, recovery is possible if you avoid the worst treatments. Also, note that the order of the treatments does not change once the simulation begins; it does not impose random losses or affect health based on prior treatment choices.\\ \\
If not otherwise specified, your response should always follow this format:\\
\textbf{<choice>Any letter in {A, B, C, D} indicates your choice of treatment</choice>
}}
\end{tcolorbox}
\caption{Prompt used for explaining the IGT task rules to the models}
\label{fig:igt-task-prompt}
\end{figure}

\begin{figure}[!h]
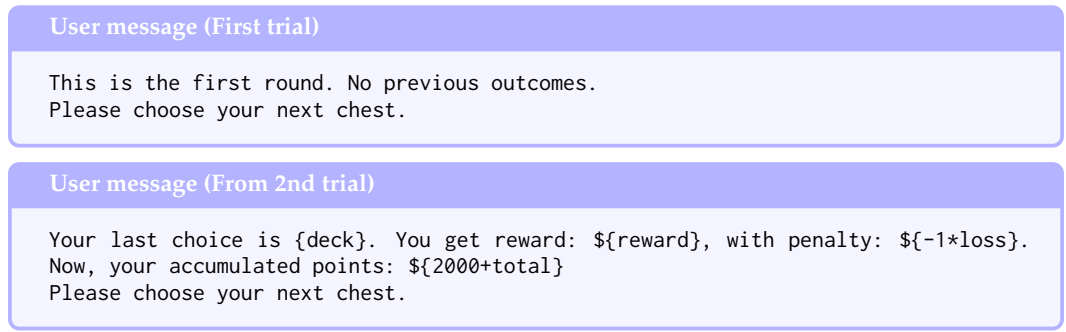


\begin{tcolorbox}[colback=blue!4, colframe=blue!30, title=\textbf{User message (First trial)}, fonttitle=\small]
\small
\texttt{This is the first round. No previous outcomes.\\
Please choose your next chest.}
\end{tcolorbox}

\begin{tcolorbox}[colback=blue!4, colframe=blue!30, title=\textbf{User message (From 2nd trial)}, fonttitle=\small]
\small
\texttt{Your last choice is \{deck\}. You get reward: \$\{reward\}, with penalty: \$\{-1*loss\}. Now, your accumulated points: \$\{2000+total\}\\
Please choose your next chest.}
\end{tcolorbox}
\caption{Prompt template used for IGT task interaction between LLM Agents and environment.}
\label{fig:game-interactive-prompt}
\end{figure}

\textbf{Interactive Prompts for IGT. } \cref{fig:game-interactive-prompt} shows basic input and responses provided by the IGT game environment as user messages. The LLM Agents perceive the consequence of their actions via these interactions.

\begin{figure}[!h]
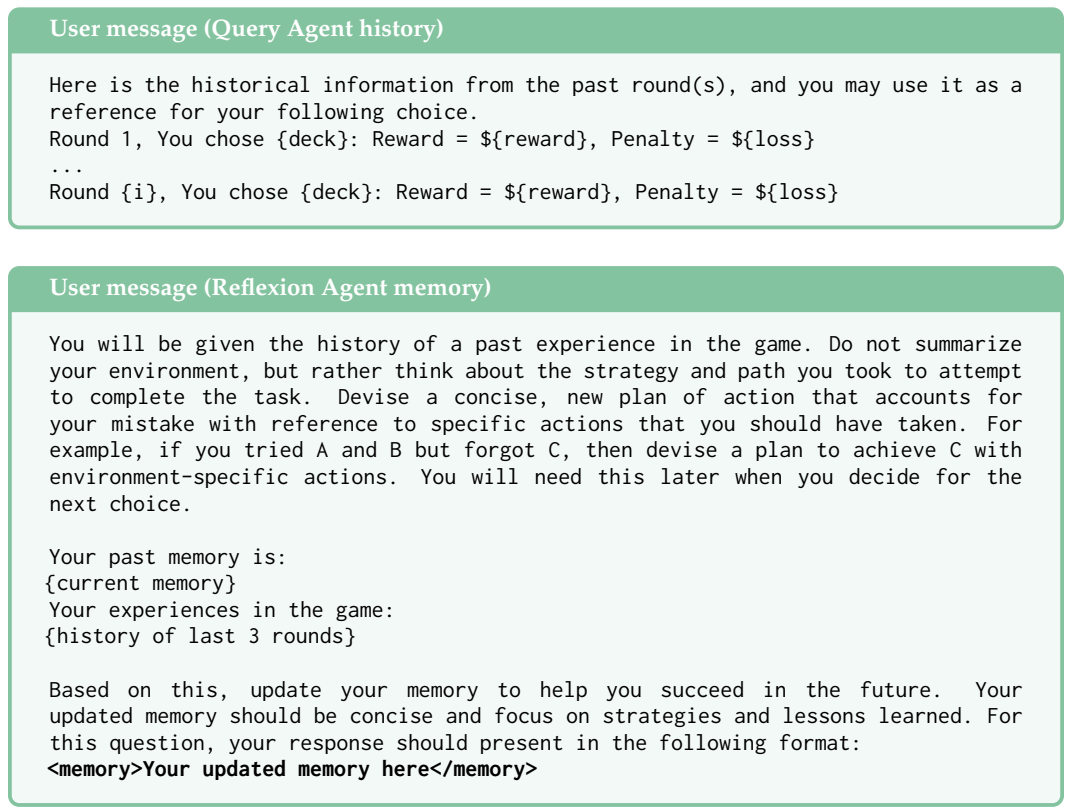

\begin{tcolorbox}[colback=ForestGreen!4, colframe=ForestGreen!50, title=\textbf{User message (Query Agent history)}, fonttitle=\small]
\small
\texttt{Here is the historical information from the past round(s), and you may use it as a reference for your following choice.\\
Round 1, You chose \{deck\}: Reward = \$\{reward\}, Penalty = \$\{loss\}\\
...\\
Round \{i\}, You chose \{deck\}: Reward = \$\{reward\}, Penalty = \$\{loss\}}
\end{tcolorbox}

\vspace{1mm}
\begin{tcolorbox}[colback=ForestGreen!4, colframe=ForestGreen!50, title=\textbf{User message (Reflexion Agent memory)}, fonttitle=\small]
\small
\texttt{You will be given the history of a past experience in the game. Do not summarize your environment, but rather think about the strategy and path you took to attempt to complete the task. Devise a concise, new plan of action that accounts for your mistake with reference to specific actions that you should have taken. For example, if you tried A and B but forgot C, then devise a plan to achieve C with environment-specific actions. You will need this later when you decide for the next choice.\\ \\
Your past memory is:\\ \{current memory\} \\ 
Your experiences in the game:\\ \{history of last 3 rounds\} \\ \\Based on this, update your memory to help you succeed in the future. Your updated memory should be concise and focus on strategies and lessons learned. For this question, your response should present in the following format: \\ \textbf{<memory>Your updated memory here</memory>}}
\end{tcolorbox}

\caption{Agent-specific prompts used in IGT task.}
\label{fig:agentic-prompt}
\end{figure}

\textbf{Agent-specific Prompts. }In addition to the shared game prompt, we use extra prompt components for agents as they require explicit access to past experience (\cref{fig:agentic-prompt}). The \textit{Query} agent receives a structured history of previous rounds, including the chosen deck and the corresponding reward and penalty, which it can use directly when selecting its next action. The \textit{Reflexion} agent receives both recent game experience and its current memory, together with an instruction to revise that memory into a concise strategy for future play. The \textit{ReAct} agent is not included in this figure because it does not use a separate agent-specific prompt template: it operates only from the standard task interaction, without an explicit retrieved-history prompt or a dedicated memory-update step.

\clearpage
\section{Robustness of Emotion Induction Validation}
\label{sec:appendix_va}
\noindent \textbf{Numerical range of Valence and Arousal.}
In our main analysis, valence is measured on $[-1,1]$ and arousal is measured on $[0,1]$. We use this asymmetric parameterization because valence is commonly treated as bipolar, while arousal is often treated as a non-negative activation dimension. This choice is not unprecedented: although some studies use the same range for both valence and arousal, such as $[-1,1]$~\citep{Mehrabian1996PAD} or $[0,1]$~\citep{mohammad-2018-obtaining}, other affective and neuroaffective studies model valence as bipolar while treating arousal as non-negative~\citep{barros2018omg, kron2013how, petrolini2020core}. In our setting, the arousal midpoint $0.5$ represents moderate arousal rather than ``zero arousal,'' so the reference center is $C=(0,0.5)$.

\begin{figure}[h]
    \centering
    \includegraphics[width=\textwidth]{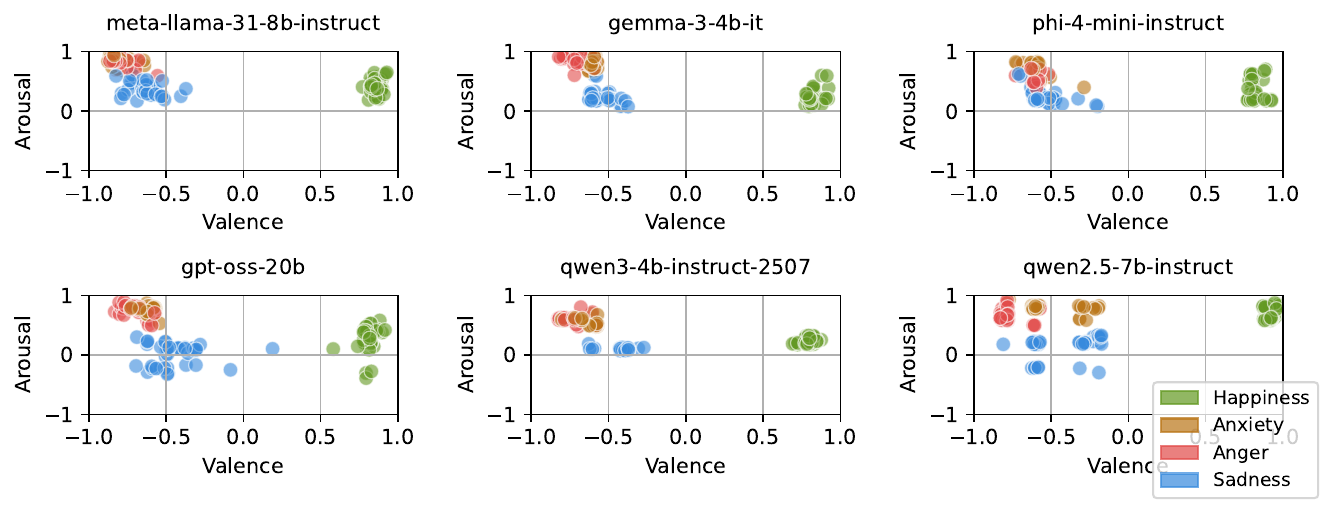}
    \caption{
    Emotion distributions where both V and A are normalized to $[-1,1]$. 
    }
    \label{fig:valence-arousal_11}
\end{figure}

\begin{figure}[h]
    \centering
    \includegraphics[width=\textwidth]{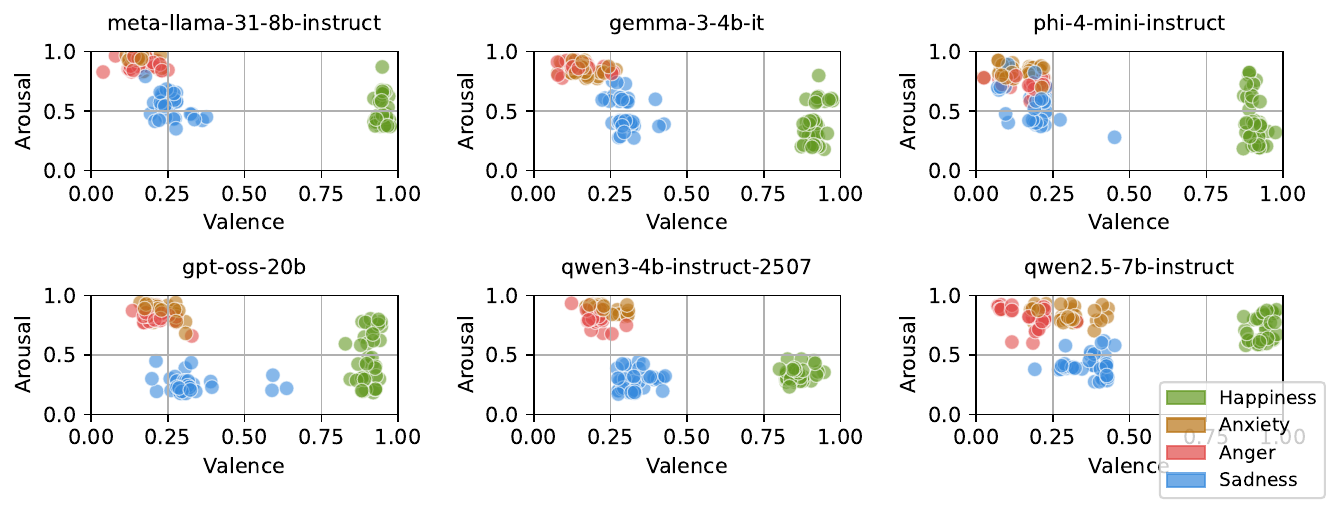}
    \caption{
    Emotion distributions where both V and A are normalized to $[0,1]$.
    }
    \label{fig:valence-arousal_01}
\end{figure}

Then came the question of whether the relative scaling of valence and arousal changes the ranking of emotions or models. To test this, we recompute the results under two uniform V--A parameterizations: $[-1,1]$ for both axes, with center $C=(0,0)$, and $[0,1]$ for both axes, with center $C=(0.5,0.5)$ (\cref{fig:valence-arousal_11,fig:valence-arousal_01}). Compared with the main visualization in \cref{fig:valence-arousal}, the relative arrangement of emotion clusters remain largely unchanged. The different V-A scales primarily introduce a linear rescaling of the coordinate system without altering the underlying affective geometry.

Table~\ref{tbl:va_scale_appendix} reports the quantitative emotion-level and model-level comparisons, which further confirms this observation quantitatively. Though the absolute metric values and rankings change by V-A parameterizations, the qualitative pattern is broadly consistnet: Anger remains a \textbf{high-separation} and \textbf{compact affective} condition across the tested scales, and the same set of top models is retained for the downstream analysis. Therefore, our qualitative conclusions are robust under V-A parameterization.

\begin{table*}[h]
\centering
\small

\begin{minipage}[t]{0.36\linewidth}
\vspace{0pt}
\caption{
Robustness of emotions under different V--A scale parameterizations.
Bold/underline indicates best/second-best values. Red background implies undesired result.
}
\label{tbl:va_scale_appendix}
\end{minipage}
\hfill
\begin{minipage}[t]{0.60\linewidth}
\vspace{0pt}
\centering
\textbf{(a) Emotion-level robustness}

\vspace{0.3em}
\resizebox{\linewidth}{!}{
\begin{tabular}{lcccccc}
\toprule
\multirow{2}{*}{\textbf{Emotion}}
& \multicolumn{2}{c}{\textbf{V$\in[-1,1]$,A$\in[0,1]$}}
& \multicolumn{2}{c}{\textbf{V$\in[-1,1]$,A$\in[-1,1]$}}
& \multicolumn{2}{c}{\textbf{V$\in[0,1]$,A$\in[0,1]$}} \\
\cmidrule(lr){2-3} \cmidrule(lr){4-5} \cmidrule(lr){6-7}
& $d(e,C)$ & $s(e)$
& $d(e,C)$ & $s(e)$
& $d(e,C)$ & $s(e)$ \\
\midrule
Anger
& \underline{0.761} & \textbf{0.132}
& \textbf{1.009} & \underline{0.410}
& \underline{0.455} & \textbf{0.087} \\
Anxiety
& 0.724 & \underline{0.150}
& \underline{0.982} & \cellcolor{red!20}0.475
& \textbf{0.475} & \textbf{0.087} \\
Sadness
& \cellcolor{red!20}0.516 & 0.183
& \cellcolor{red!20}0.535 & \textbf{0.220}
& \cellcolor{red!20}0.220 & \underline{0.179} \\
Happiness
& \textbf{0.849} & \cellcolor{red!20}0.210
& 0.910 & 0.456
& 0.411 & \cellcolor{red!20}0.199 \\
\bottomrule
\end{tabular}
}
\end{minipage}

\vspace{0.7em}

\textbf{(b) Model-level robustness of anger emotion in different V-A scales.}

\vspace{0.3em}
\resizebox{\linewidth}{!}{
\begin{tabular}{lccccccccc}
\toprule
\multirow{2}{*}{\textbf{Model}}
& \multicolumn{3}{c}{\textbf{V$\in[-1,1]$, A$\in[0,1]$}}
& \multicolumn{3}{c}{\textbf{V$\in[-1,1]$, A$\in[-1,1]$}}
& \multicolumn{3}{c}{\textbf{V$\in[0,1]$, A$\in[0,1]$}} \\
\cmidrule(lr){2-4} \cmidrule(lr){5-7} \cmidrule(lr){8-10}
& $d(A,S)$ & $d(A,C)$ & $s(A)$
& $d(A,S)$ & $d(A,C)$ & $s(A)$
& $d(A,S)$ & $d(A,C)$ & $s(A)$ \\
\midrule
Qwen3-4B
& \textbf{0.30} & \underline{0.81} & 0.10
& \textbf{0.30} & 0.76 & \textbf{0.08}
& \textbf{0.27} & 0.38 & \textbf{0.05} \\
Llama-3.1-8B
& \underline{0.20} & \textbf{0.90} & \textbf{0.07}
& 0.10 & \textbf{0.87} & 0.10
& 0.05 & \textbf{0.42} & 0.06 \\
GPT-oss-20b
& \underline{0.20} & \underline{0.76} & 0.10
& \underline{0.22} & 0.72 & 0.12
& \underline{0.26} & \underline{0.40} & \textbf{0.05} \\
Qwen2.5-7B
& 0.10 & 0.70 & 0.11
& \textbf{0.30} & \underline{0.76} & 0.14
& 0.10 & 0.37 & 0.10 \\
Gemma3-4B
& \cellcolor{red!20}0.00 & 0.77 & 0.09
& \cellcolor{red!20}0.00 & 0.75 & 0.11
& 0.05 & 0.39 & 0.06 \\
Phi-4-Mini
& \cellcolor{red!20}0.00 & \cellcolor{red!20}0.63 & 0.08
& \cellcolor{red!20}0.00 & \cellcolor{red!20}0.61 & 0.11
& \cellcolor{red!20}0.00 & \cellcolor{red!20}0.28 & 0.08 \\
\bottomrule
\end{tabular}
}

\end{table*}

\begin{wraptable}{r}{0.5\linewidth}
\centering
\small
\caption{Discrete-emotion validation. Mismatch is the percentage of generated vignettes classified as a non-target emotion.}
\label{tab:discrete_emotion_validation}
\resizebox{\linewidth}{!}{
\begin{tabular}{clc}
\toprule
\textbf{Rank} & \textbf{Model} & \textbf{Mismatch (\%)} \\
\midrule
1 & Gemma3-4B & 0.000 \\
2 & Qwen2.5-7B-Instruct & 1.875 \\
3 & Phi-4-Mini-Instruct & 3.125 \\
4 & Llama-3.1-8B-Instruct & 5.625 \\
5 & Qwen3-4B-Instruct-2507 & 6.250 \\
6 & GPT-oss-20b & 7.500 \\
\bottomrule
\end{tabular}
}
\end{wraptable}

\paragraph{Discrete-emotion validation.}
To further validate whether the generated vignettes express the intended discrete emotions, we add two checks beyond the V--A separability analysis. \textbf{First}, in a separate forward-pass prompt, we ask LLM annotators to classify the emotion expressed in each vignette among the target categories, using the prompt in~\cref{fig:va-prompt-classification}. The LLM-based validation shows high agreement with the intended labels, reaching 96\% accuracy overall. Table~\ref{tab:discrete_emotion_validation} reports the mismatch rate by model. \textbf{Second}, we conduct a human annotation check on 80 randomly shuffled vignettes, with 20 vignettes from each target emotion. Three human annotators with English proficiency ranging from CEFR B2 to C2 are tasked with \textit{Classify the emotional content of the text below into one of four categories: ’anxiety’, ’sadness’, ’anger’, or ’happiness’.}. The participants correctly identified 73, 76, and 79 out of 80 vignettes, respectively, giving an average accuracy of 95\%. While high accuracy is partly expected because the vignettes are generated from emotion-targeted prompts, these results strengthen the validity of our method.

\newpage
\section{Ablation Study}
\label{sec:ablation}
We define \textit{Profit\%} as the percentage change in total money relative to the initial budget at the end of the game:
$\textit{Profit\%} = 100 \times \frac{\textit{total\_money} - 2000}{2000}$.

We report this metric in addition to IGT score because IGT score only counts the relative number of advantageous versus disadvantageous choices, whereas \textit{Profit\%} captures the actual accumulated monetary outcome. As a result, two runs with similar IGT scores may still differ in final profit. For each emotion comparison, we test whether the mean metric differs between the anger and neutral conditions using a two-sided $t$-test over repeated runs. 

\begin{table}[ht]
\centering
\small
\setlength{\tabcolsep}{4.5pt}
\renewcommand{\arraystretch}{0.88}
\caption{Effects of agent design and anger induction across models. \textbf{IGT score@1/@2} denote block-wise net scores for Blocks 1/2, and \textbf{IGT score} denotes the 40-round net score. $\mathbf{\Delta}$ is anger $-$ neutral, tested by a two-sided paired $t$-test over matched seeds and deck orders. Anger effects are mostly non-significant and inconsistent across settings.} 
\label{tab:ablation-agents-models}
\begin{tabular}{llcccc}
\toprule
\multicolumn{6}{c}{\textbf{GPT-oss-20b}} \\
\midrule
Agent & Emotion & IGT score@1$\uparrow$ & IGT score@2$\uparrow$ & IGT score$\uparrow$ & Profit\%$\uparrow$ \\
\midrule
\multirow{3}{*}{ReAct}
& none  & -2.22 $\pm$ 1.84 & 4.11 $\pm$ 1.22 & 1.89 $\pm$ 2.63 & -4.38 $\pm$ 6.84 \\
& anger & -2.56 $\pm$ 1.79 & 6.11 $\pm$ 1.50 & 3.56 $\pm$ 2.68 & 1.18 $\pm$ 7.06 \\
 & $\Delta$ & -0.33 (ns) & +2.00 (ns) & +1.67 (ns) & +5.56 (ns) \\
\hline
\multirow{3}{*}{Query}
& none  & 1.44 $\pm$ 2.46 & 7.78 $\pm$ 2.13 & 9.22 $\pm$ 3.84 & 18.41 $\pm$ 9.21 \\
& anger & 0.44 $\pm$ 2.48 & 8.78 $\pm$ 2.05 & 9.22 $\pm$ 3.57 & 23.13 $\pm$ 8.67 \\
 & $\Delta$ & -1.00 (ns) & +1.00 (ns) & 0.00 (ns) & +4.72 (ns) \\
\hline
\multirow{3}{*}{Reflexion}
& none  & -4.56 $\pm$ 2.11 & 12.22 $\pm$ 2.62 & 7.67 $\pm$ 2.98 & 25.69 $\pm$ 7.15 \\
& anger & -2.11 $\pm$ 2.32 & 9.67 $\pm$ 2.73 & 7.56 $\pm$ 2.71 & 22.22 $\pm$ 7.29 \\
 & $\Delta$ & +2.44 (ns) & -2.56 (ns) & -0.11 (ns) & -3.47 (ns) \\

\toprule
\multicolumn{6}{c}{\textbf{Qwen3-4B-Instruct-2507}} \\
\midrule
Agent & Emotion & IGT score@1$\uparrow$ & IGT score@2$\uparrow$ & IGT score$\uparrow$ & Profit\%$\uparrow$ \\
\midrule
\multirow{3}{*}{ReAct}
& none  & 1.22 $\pm$ 1.91 & 3.56 $\pm$ 3.39 & 4.78 $\pm$ 5.27 & 17.85 $\pm$ 11.03 \\
& anger & -0.56 $\pm$ 1.73 & 1.89 $\pm$ 2.90& 1.33 $\pm$ 4.58 & -0.35 $\pm$ 12.89 \\
 & $\Delta$ & -1.78 * & -1.67 (ns) & -3.44 (ns) & -18.19 * \\
\hline
\multirow{3}{*}{Query}
& none  & -2.11 $\pm$ 1.25 & -2.89 $\pm$ 1.81 & -5.00 $\pm$ 2.02 & -0.83 $\pm$ 4.40 \\
& anger & -2.67 $\pm$ 1.64 & -3.11 $\pm$ 3.75 & -5.78 $\pm$ 2.97 & -5.21 $\pm$ 10.35 \\
 & $\Delta$ & -0.56 (ns) & -0.22 (ns)  & -0.78 (ns) & -4.38 (ns) \\
\hline
\multirow{3}{*}{Reflexion}
& none  & -2.11 $\pm$ 2.48 & 7.22 $\pm$ 3.78 & 5.11 $\pm$ 4.83 & 13.82 $\pm$ 8.07 \\
& anger & 0.89 $\pm$ 2.60 & 3.78 $\pm$ 4.42 & 4.66 $\pm$ 5.89 & 10.28 $\pm$ 9.81 \\
 & $\Delta$ & +3.00 (ns) & -3.44 (ns) & -0.44 (ns) & -3.54 (ns) \\
\toprule
\multicolumn{6}{c}{\textbf{Qwen2.5-7B-Instruct}} \\
\midrule
Agent & Emotion & IGT score@1$\uparrow$ & IGT score@2$\uparrow$ &  IGT score$\uparrow$ & Profit\%$\uparrow$ \\
\midrule
\multirow{3}{*}{ReAct}
& none  & 0.00 $\pm$ 0.00 & 0.00 $\pm$ 0.00 & 0.00 $\pm$ 0.00 & 0.00 $\pm$ 0.00 \\
& anger & -0.33 $\pm$ 0.33 & 0.44 $\pm$ 0.35 & 0.11 $\pm$ 0.25 & -0.42 $\pm$ 0.42 \\
 & $\Delta$ & -0.33 (ns) & +0.44 (ns) & +0.11 (ns) & -0.42 (ns) \\
\hline
\multirow{3}{*}{Query}
& none  & -1.44 $\pm$ 1.25 & -2.67 $\pm$ 3.07 & -4.11 $\pm$ 4.12 & 3.54 $\pm$ 10.12 \\
& anger & -3.11 $\pm$ 1.46 & -8.56 $\pm$ 2.35 & -11.67 $\pm$ 2.97 & -9.65 $\pm$ 7.51 \\
 & $\Delta$ & -1.67 (ns) & -5.89 (ns) & -7.56 (ns) & -13.19 (ns) \\
\hline
\multirow{3}{*}{Reflexion}
& none  & 0.00 $\pm$ 2.21 & 1.78 $\pm$ 3.38 & 1.78 $\pm$ 4.80 & 17.57 $\pm$ 11.39 \\
& anger & -0.56 $\pm$ 2.48 & 1.22 $\pm$ 3.98 & 0.67 $\pm$ 5.33 & 11.11 $\pm$ 9.77 \\
& $\Delta$ & -0.56 (ns) & -0.56 (ns) & -1.11 (ns) & -6.46 (ns) \\
\toprule
\multicolumn{6}{c}{\textbf{Llama3.1-8B-Instruct}} \\
\midrule
Agent & Emotion & IGT score@1$\uparrow$ & IGT score@2$\uparrow$ &  IGT score$\uparrow$ & Profit\%$\uparrow$ \\
\midrule
\multirow{3}{*}{ReAct}
& none  & -2.11 $\pm$ 0.75 & -0.56 $\pm$ 1.12 & -2.67 $\pm$ 1.39 & 4.44 $\pm$ 6.00 \\
& anger & -2.89 $\pm$ 1.20 & -3.78 $\pm$ 1.91 & -6.67 $\pm$ 2.43 & -0.83 $\pm$ 5.82 \\
 & $\Delta$ & -0.78 (ns) & -3.22 (ns) & -4.00 (ns) & -5.27 (ns) \\
\hline
\multirow{3}{*}{Query}
& none  & -1.33 $\pm$ 1.65 & -2.67 $\pm$ 1.62 & -4.00 $\pm$ 3.11 & 6.25 $\pm$ 9.51 \\
& anger & -3.11 $\pm$ 1.62 & -5.22 $\pm$ 1.16 & -8.33 $\pm$ 2.46 & -5.90 $\pm$ 8.40 \\
 & $\Delta$ & -1.78 (ns) & -2.56 (ns) & -4.33 (ns) & -12.15 (ns) \\
\hline
\multirow{3}{*}{Reflexion}
& none  & 1.44 $\pm$ 1.28 & 3.56 $\pm$ 2.08 & 5.00 $\pm$ 2.37 & 32.63 $\pm$ 8.06 \\
& anger & -1.78 $\pm$ 1.60 & 2.00 $\pm$ 1.68 & 0.22 $\pm$ 2.10 & 13.06 $\pm$ 6.41 \\
& $\Delta$ & -3.22 (ns) & -1.56 (ns) & -4.78 (ns) & -19.58 (ns) \\
\bottomrule
\end{tabular}
\end{table}

\cref{tab:ablation-agents-models} shows the performance of different agent designs across models under neutral and anger conditions. While some mean differences can be observed across agents, blocks, and metrics, these effects are inconsistent in both magnitude and direction. Moreover, nearly all comparisons are statistically non-significant, suggesting that these variations are not robust. This means that differences across emotion conditions are not consistent across agent–model combinations, indicating no stable effect of emotion induction.

Since ~\cref{tab:ablation-agents-models} contains mostly non-significant neutral-vs-anger differences, we further examine whether these null results reflect a lack of reliable average effects or simply insufficient sensitivity. Non-significance alone should not be interpreted as evidence of no effect. Therefore, we report the paired neutral-vs-anger $p$-values in ~\cref{tab:neutral_anger_pvalues} as supplementary to ~\cref{tab:ablation-agents-models}; and the 80\% minimum detectable effects (MDEs) in ~\cref{tab:mde_single_block}. MDE is the smallest detectable anger--neutral difference in block-wise net score with 80\% power at $\alpha=0.05$; smaller is more sensitive.

\begin{table*}[h]
\centering
\small
\caption{
Paired neutral-vs-anger $t$-test $p$-values across model-agent configurations. Only 2 values reaches significance, suggesting no stable anger effect. *:$p<0.05$.
}
\label{tab:neutral_anger_pvalues}
\resizebox{\linewidth}{!}{
\begin{tabular}{llcccc}
\toprule
\textbf{Model} & \textbf{Agent} 
& \textbf{IGT score@1} 
& \textbf{IGT score@2} 
& \textbf{IGT score} 
& \textbf{Profit\%} \\
\midrule
\multirow{3}{*}{GPT-oss-20b}
& ReAct     & 0.886 & 0.177 & 0.568 & 0.447 \\
& Query     & 0.386 & 0.690 & 1.000 & 0.521 \\
& Reflexion & 0.295 & 0.370 & 0.980 & 0.750 \\
\midrule
\multirow{3}{*}{Qwen3-4B-Instruct-2507}
& ReAct     & \textbf{0.038*} & 0.456 & 0.247 & \textbf{0.017*} \\
& Query     & 0.714 & 0.874 & 0.701 & 0.618 \\
& Reflexion & 0.375 & 0.534 & 0.951 & 0.765 \\
\midrule
\multirow{3}{*}{Qwen2.5-7B-Instruct}
& ReAct     & 0.331 & 0.215 & 0.790 & 0.331 \\
& Query     & 0.205 & 0.108 & 0.093 & 0.226 \\
& Reflexion & 0.826 & 0.917 & 0.855 & 0.603 \\
\midrule
\multirow{3}{*}{Llama3.1-8B-Instruct}
& ReAct     & 0.591 & 0.150 & 0.147 & 0.370 \\
& Query     & 0.265 & 0.186 & 0.153 & 0.274 \\
& Reflexion & 0.137 & 0.587 & 0.156 & 0.052 \\
\bottomrule
\end{tabular}
}
\end{table*}

Results show that only a small number of comparisons reach statistical significance (ReAct $\times$ Qwen3-4B-Instruct-2507 for IGT score@1 and Profit\%), which supports the conclusion that anger does not produce a reliable average shift across model--agent configurations. At the same time, the MDE analysis clarifies the scale of effects that our experiments can rule out. For single-block (20-round) IGT scores, one shifted choice from a disadvantageous deck to an advantageous deck changes the net score by 2 points. Thus, MDE values around 2-4 correspond to roughly one to two shifted choices within a 20-round block, while MDE values around 6-8 correspond to roughly three to four shifted choices. As shown in~\cref{tab:mde_single_block}, Query and ReAct agents are sensitive to relatively small block-level changes in several models, whereas Reflexion has larger MDEs and should therefore be interpreted more cautiously.

\begin{table}[h]
\centering
\small
\caption{
80\% minimum detectable effects (MDEs) for 20-round block IGT scores under paired $t$-tests. Smaller values imply higher sensitivity.
It is noticeable that Query/ReAct are more sensitive than Reflexion agents.
}
\label{tab:mde_single_block}
\resizebox{\linewidth}{!}{
\begin{tabular}{lcccc}
\toprule
\textbf{Agent} 
& \textbf{GPT-oss-20b} 
& \textbf{Qwen3-4B-Instruct-2507} 
& \textbf{Qwen2.5-7B-Instruct} 
& \textbf{Llama3.1-8B-Instruct} \\
\midrule
Query     & 3.339 & 4.427 & 3.762 & 4.583 \\
ReAct     & 4.218 & 2.349 & 1.223 & 4.222 \\
Reflexion & 6.722 & 7.995 & 7.381 & 6.140 \\
\bottomrule
\end{tabular}
}
\end{table}

\clearpage
\begin{figure*}
    \centering
    \includegraphics[width=0.46\linewidth]{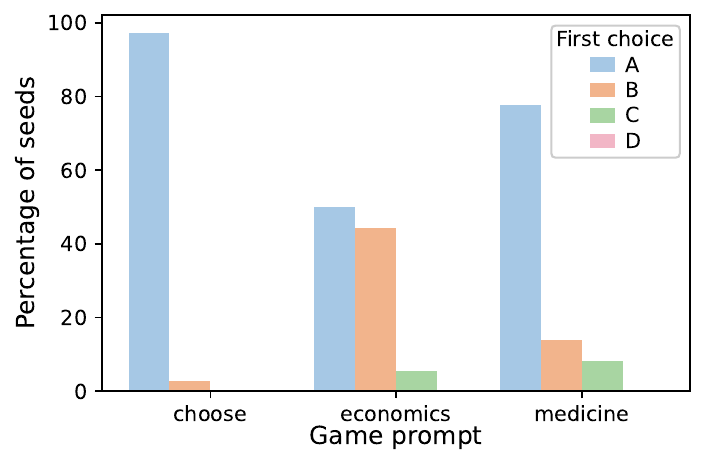}\hfill
    \includegraphics[width=0.53\linewidth]{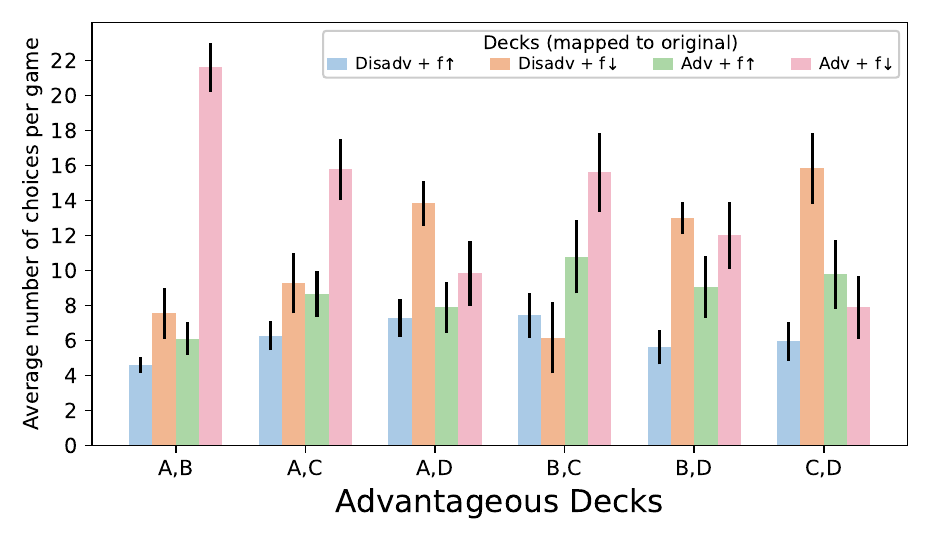}
    \vspace{0.2em}
    {\small (a) GPT-oss-20b \par}
    \vspace{0.6em}
    \includegraphics[width=0.46\linewidth]{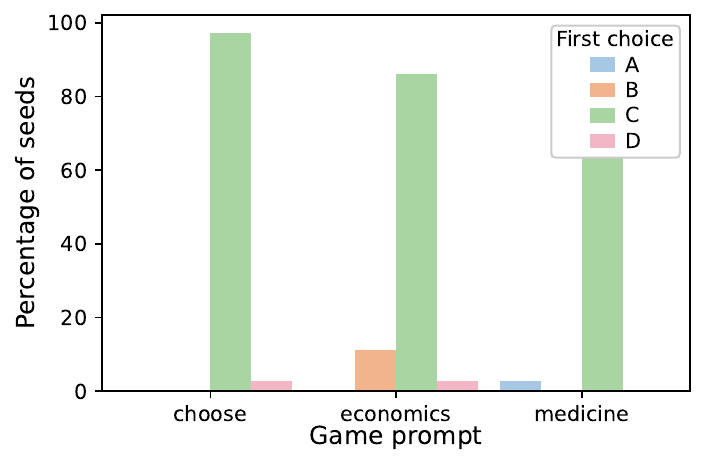}\hfill
    \includegraphics[width=0.53\linewidth]{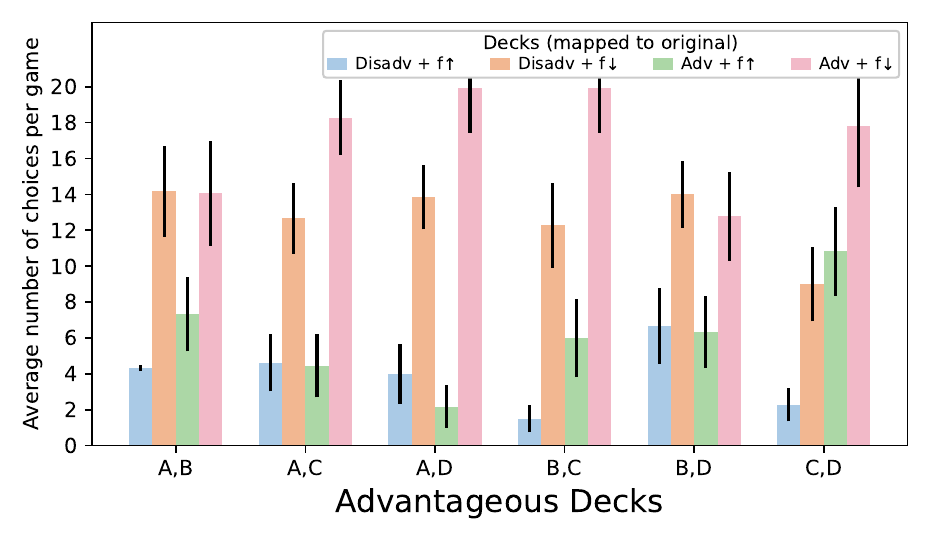}
    \vspace{0.2em}
    {\small (b) Qwen3-4B-Instruct-2507 \par}
    \vspace{0.6em}
    \includegraphics[width=0.46\linewidth]{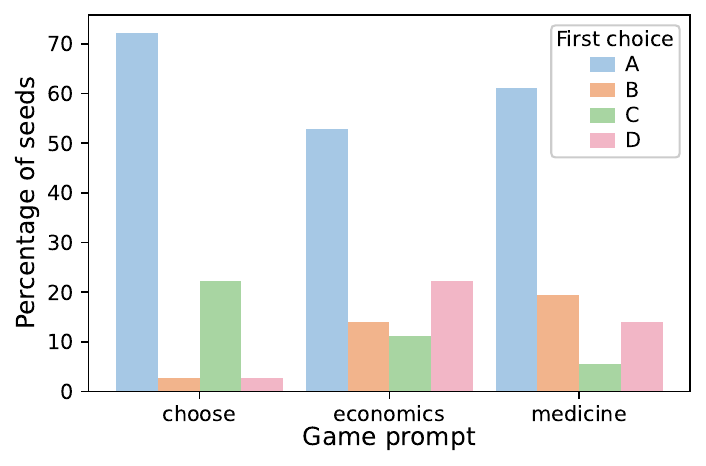}\hfill
    \includegraphics[width=0.53\linewidth]{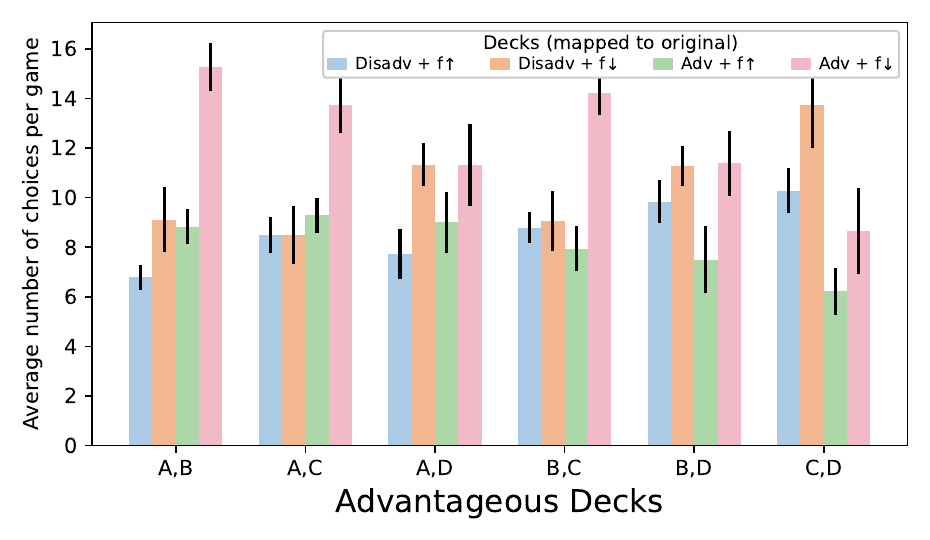}
    \vspace{0.2em}
    {\small (c) Llama3.1-8B-Instruct \par}
    \caption{
    \textbf{(Left: Initial Bias)} Different models have different biases at the first choice. \textbf{(Right: Game Bias)} Across advantageous-deck conditions, the agent’s selections are systematically skewed toward decks with less frequent losses. The order of decks changes the odds of the advantageous decks to be chosen.}
    \label{fig:bias_and_advdecks}
\end{figure*}
\section{LLMs' Biased Initial Choices and Consequences on Information Bias.}
\label{sec:app_bias}
As observed in multiple prior works \citep{zhenglarge, pezeshkpour2024large}, LLMs present a bias to how choices are present in prompts. In our case, the model is heavily biased towards choosing some decks/letters in its first round (\cref{fig:bias_and_advdecks} (left)), which makes the original fixed deck ordering problematic: any early preference may reflect positional bias rather than learned preference from feedback. As mentioned in \cref{sec:emo_ind}, we periodically shuffled the decks, so that the results would not systematically correspond to the first option. 

The results in \cref{fig:bias_and_advdecks} (right) indicate that deck shuffling mitigates, but does not remove, prompt-order bias. Importantly, the agent still selects \textit{D} most frequently on average, suggesting that it can learn to favor an advantageous deck despite randomized label-to-deck mappings. However, residual asymmetries remain: in configurations where \textit{A} is not advantageous, the agent often shifts toward disadvantageous deck \textit{B}, which also has low loss frequency. This pattern suggests that learning is preserved under shuffling, but choice behavior is still partly shaped by how options are presented.

\begin{wrapfigure}{l}{0.5\textwidth}
    \centering
    \vspace{-0.2in}
    \includegraphics[width=\linewidth]{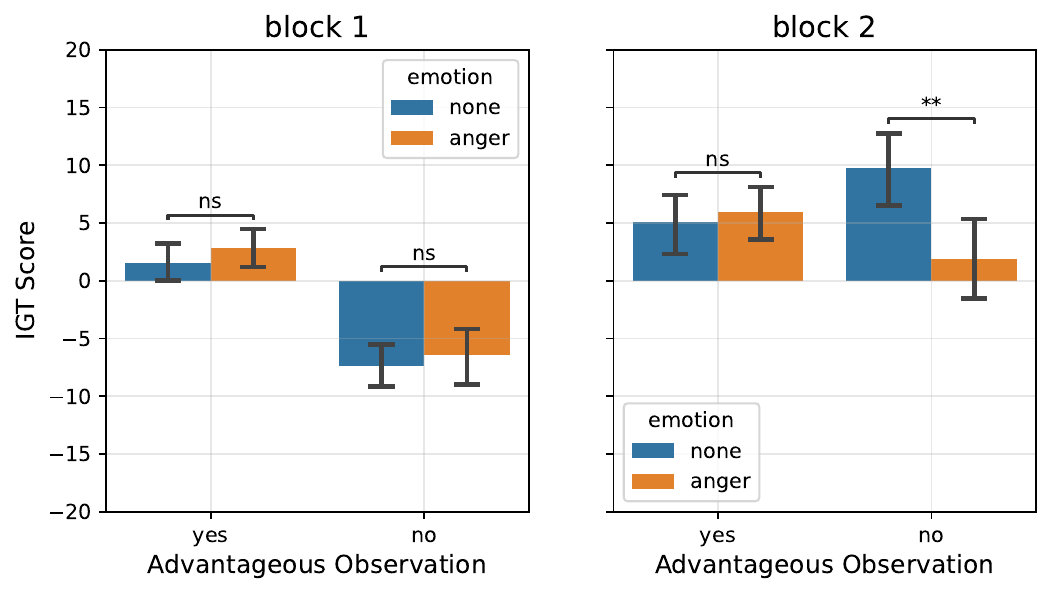}
    \caption{Emotion causes different decision patterns in the 2nd block under impaired early observation.}
    \label{fig:igt-choose-2block_anger}
    \vspace{-0.1in}
\end{wrapfigure}

Based on this observation, we define \textit{Advantageous Observation} as selecting an advantageous deck by chance on the first round, given the model’s strong tendency to choose one of the four options. For example, if the model chooses \textit{'A'} on its first round and the advantageous decks are \textit{B} and \textit{C}, it will count as a \textit{'no'}.

\textbf{Anger Emotion Leads to ``Gut-Feeling'' Decision Making.}
\cref{fig:igt-choose-2block_anger} shows that the effect of emotion depends on whether the advantageous deck is observed, and that this difference emerges mainly in block 2 rather than block 1. In block 1, IGT scores are low and the differences between neutral and anger conditions are not significant in either observation setting, suggesting that both agents are still in an early exploratory phase. In block 2, however, the pattern becomes asymmetric. When the advantageous deck is observed, performance remains similar under neutral and anger prompting, with no significant difference between the two conditions. In contrast, when the advantageous deck is not observed, the neutral agent achieves clearly higher IGT scores, whereas the anger-conditioned agent performs substantially worse, and this gap is significant. Thus, anger does not broadly impair decision making across all settings; rather, anger-conditioned agents adapt more weakly under ambiguous conditions.

\textbf{Emotion Mediate How Agents Explore/Exploit Games.}
The moderation results lead to a new question: why induced anger makes the LLM less sensitive to the outcome of disadvatageous decks.
We hypothesize that the moderation is associated with the exploitation/exploration balance of the agent.
To examine the exploitation/exploration strategies of agents,
we fit a simple choice model to its sequential decisions. Before each turn, we reconstructed the agent's belief state over the four decks using only its past observations. For each deck $d$, we calculate an estimated value $\mu_d$ as the mean of previously observed net outcomes from that deck, and an uncertainty term $\sigma_d = 1/\sqrt{n_d+1}$, where $n_d$ is the number of times the deck had been sampled so far. 
We then modeled the probability of choosing deck $d$ with a softmax policy
\begin{equation}
    P(d_t = d) \propto \exp(\beta \mu_{t,d} + \gamma \sigma_{t,d}) \label{eq:exploration}
\end{equation}
where $\beta$ captures sensitivity to expected reward and $\gamma$ captures sensitivity to uncertainty. 
Parameters were estimated by maximizing the log-likelihood of the observed choices. Under this formulation, a larger $\beta$ indicates that the agent's choices are more strongly guided by its learned reward estimates, whereas a larger $\gamma$ indicates greater sensitivity to uncertainty.

\begin{wrapfigure}{r}{0.5\textwidth}
    \centering
    \includegraphics[width=\linewidth]{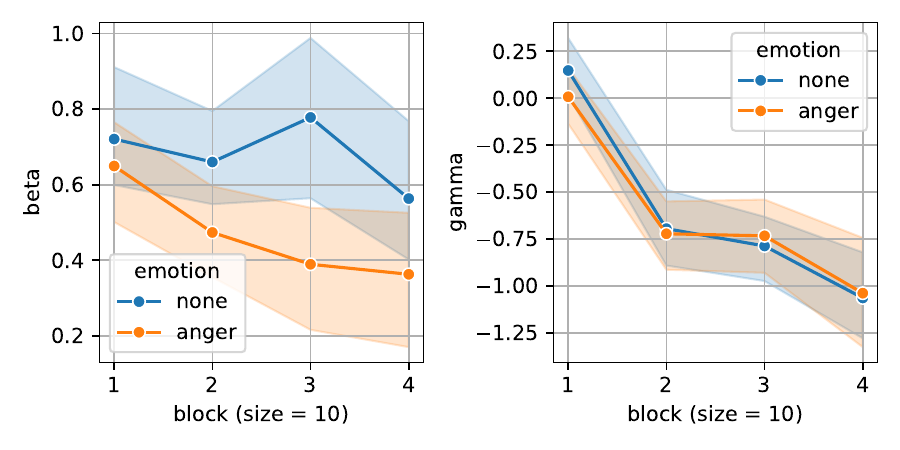}
    \caption{Anger emotion induces less exploitation ($\beta\downarrow$) on Reflexion $\times$ GPT-oss-20b.} 
    \label{fig:igt-choose-anger-beta-gamma}
\end{wrapfigure}

In \cref{fig:igt-choose-anger-beta-gamma}, we plot the estimated $\beta$ and $\gamma$ parameters by a \emph{10-round-sized} block for the neutral and anger conditions.
Across blocks, the anger condition generally exhibited lower $\beta$ estimates than the neutral condition at the 3rd block, indicating a weaker tendency to exploit learned value estimates.

Meanwhile, under both emotions, exploration tends to decrease across trials. 
That can be attributed to the fact that $\sigma_d$ diminishes by trials and there are less new trials to explore.
Though with similar trend, we can see that a higher $\gamma$ was observed in the anger condition at rounds 20 to 40, suggesting a possible increase in uncertainty sensitivity.

\end{document}